\documentclass[10pt,journal,compsoc]{IEEEtran}
\ifCLASSOPTIONcompsoc
  \usepackage[nocompress]{cite}
\else
  \usepackage{cite}
\fi
\usepackage{hyperref}
\usepackage{epsfig}
\usepackage{amsmath,amssymb,amsfonts,mathrsfs,bm,bbm}
\usepackage{amssymb}
\usepackage{stfloats}
\usepackage{booktabs}
\usepackage{array}
\usepackage{algorithm}
\usepackage{algorithmicx}
\usepackage{algpseudocode}
\usepackage{amssymb}
\usepackage{bbding}
\usepackage{wasysym}
\usepackage{pifont}
\usepackage{graphicx}
\usepackage{subfigure}
\usepackage{graphics}
\usepackage{multirow}
\newcommand{\xmark}{\ding{55}}%
\ifCLASSINFOpdf
\else
\fi
\begin{document}
%
\title{Disentangling Identity and Pose for Facial Expression Recognition}
%
%
%
%

\author{Jing~Jiang and
        Weihong~Deng,~\IEEEmembership{Member,~IEEE}
\IEEEcompsocitemizethanks{\IEEEcompsocthanksitem The authors are with School of Artificial Intelligence, Beijing University of Posts and Telecommunications, Beijing, 100876, China. Weihong Deng is the corresponding author.
E-mail:\{jiangjing1998,whdeng\} @bupt.edu.cn.}
}

%
%

\markboth{Journal of \LaTeX\ Class Files,~Vol.~14, No.~8, August~2015}%
{Shell \MakeLowercase{\textit{et al.}}: Bare Demo of IEEEtran.cls for Computer Society Journals}
%



\IEEEtitleabstractindextext{%
\begin{abstract}
Facial expression recognition (FER) is a challenging problem because the expression component is always entangled with other irrelevant factors, such as identity and head pose. In this work, we propose an identity and pose disentangled facial expression recognition (IPD-FER) model to learn more discriminative feature representation. We regard the holistic facial representation as the combination of identity, pose and expression. These three components are encoded with different encoders. For identity encoder, a well pre-trained face recognition model is utilized and fixed during training, which alleviates the restriction on specific expression training data in previous works and makes the disentanglement practicable on in-the-wild datasets. At the same time, the pose and expression encoder are optimized with corresponding labels. Combining identity and pose feature, a neutral face of input individual should be generated by the decoder. When expression feature is added, the input image should be reconstructed. By comparing the difference between synthesized neutral and expressional images of the same individual, the expression component is further disentangled from identity and pose. Experimental results verify the effectiveness of our method on both lab-controlled and in-the-wild databases and we achieve state-of-the-art recognition performance.
\end{abstract}

}

\maketitle

\IEEEdisplaynontitleabstractindextext

%
\IEEEpeerreviewmaketitle

\IEEEraisesectionheading{\section{Introduction}\label{sec:introduction}}
Facial expression recognition (FER) attracts much attention in computer vision since it's significant for many applications such as human-computer interaction, social interaction analysis, and medical treatments. Despite much progress has made in FER with the development of deep learning, there are still many challenges\cite{li2020deep}. One significant difficulty comes from the entanglement of expression and identity. The variations between different expressions of the same person is subtler than the discrepancy of different individuals. The learned expressional representation remains identity-dominated information without sufficient training data, which makes the model fail to generalize well on unseen identities. 
\begin{figure}[http]
  \centering
  \includegraphics[scale=0.42]{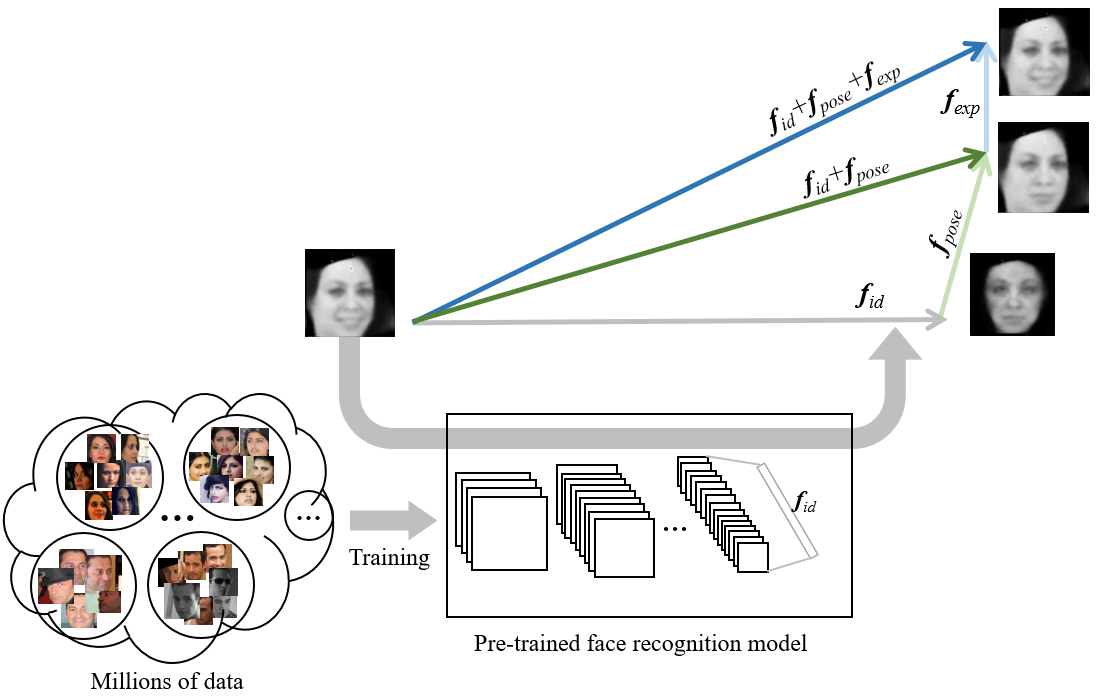}
  \caption{We model the holistic facial feature as combination of identity feature $f_{id}$, pose feature $f_{pose}$ and expression feature $f_{exp}$. A pre-trained face recognition model is utilized to extract identity feature, which is expression-invariant and pose-invariant with the help of large-scale face recognition training data. Expression labels are provided and pose labels are easy to acquire, they can serve as supervision for obtaining corresponding features.}
  \label{intro}
\end{figure}
To alleviate the adverse effect caused by identity, researchers have tried to learn person-independent representations for FER. Considering a facial image can be regarded as the combination of identity and expression components, previous works \cite{yang2018facial}, \cite{liu2017adaptive}, \cite{bai2019disentangled} compare the difference between the query image and corresponding neutral one so that the expression component is reserved while the identity information is removed. However, these disentanglement algorithms rely deeply on specific databases that the neutral-expressional image pairs are available for each individual, which are hard to collect in the wild. \cite{liu2019hard} extends \cite{liu2017adaptive} and tries to synthesize neutral faces before the disentanglement algorithm is conducted. These approaches perform well on lab-controlled databases, but for more complex in-the-wild facial expression recognition, their strengths may be weakened due to higher diversity of query images and lack of the required image pairs. Expression transferring based methods disentangle identity and expression from a new perspective. TDGAN \cite{xie2020facial} and TER-GAN \cite{ali2021facial} use two branches of encoder to extract identity and expression features, respectively. Two facial images from two individuals A and B are fed into network, then the decoder should construct a new image which keeps the same identity with person A but the same expression with person B, with features combined from the two encoders. However, they also don't perform well on in-the-wild data. Recently,\cite{cai2021identity} aims to generate an "average" image calculated by multiple individuals for each expression and remove the identity information. But for spontaneous expressions in not-controlled environment, it may lose some diversity and complexity of data due to the average operation. 

Another interference factor for FER is head pose. A wide range of approaches have been proposed to achieve pose-robust FER \cite{rudovic2012coupled,eleftheriadis2014discriminative,zhang2016deep,wu2017locality}. But they rely on multi-view datasets and fail to generalize on in-the-wild data. \cite{wang2020region} tries to address real-world pose-robust FER problem by capturing the most important facial regions. To cope with both identity and pose in not-controlled environment, an identity-robust and pose-robust model \cite{wang2019identity} is trained by adversarial learning. However, the identity labels are required, which is not available for existing in-the-wild databases. 


Considering the limitation of previous methods, we aim to achieve more robust FER in the wild through disentangling expression from both identity and head pose. Firstly, we derive identity-irrelevant expression representation by comparing the neutral and expressional version of the same person, just as previous works \cite{yang2018facial}, \cite{liu2017adaptive}, \cite{bai2019disentangled}. Differently, we utilize a pre-trained face recognition model to serve as neutral reference. As shown in Fig. \ref{intro}, with millions of training data and well-designed loss functions, the well trained face recognition model is able to extract discriminative identity feature which is irrelevant to expression and head pose \cite{deng2019arcface,wang2021deep}. Thus, the pre-trained model can serve as identity reference, which is approximatively neutral and frontal. Because the pre-trained model can be obtained by downloading from Internet or training with face recognition datasets, it avoids the requirement of neutral faces or identity labels for FER data. However, the query faces are not always frontal for in-the-wild FER problem. To realize both identity and pose invariant FER, the neutral reference should be with not only the same identity but also the same head pose as query image. Therefore, we model the holistic facial feature as the vector sum of identity feature $f_{id}$, pose feature $f_{pose}$ and expression feature $f_{exp}$, as shown in Fig. \ref{intro}. For the query expressional image, it should be encoded with combination of above three feature components. As for its neutral reference, it only contains identity and pose information.

To learn desired discriminative expression feature, we propose an identity and pose disentangled facial expression recognition model, named IPD-FER. The model is based on Generative Adversarial Network (GAN), as shown in Fig. \ref{method}. The generator $G$ consists of three encoders $E_{id}$, $E_{exp}$ and $E_{pose}$, and one decoder $G_{dec}$. The overall facial representation in feature space $f_{i-p-e}$ is the combination of three independent components, which are identity feature $f_{id}$, pose feature $f_{pose}$, and expression feature $f_{exp}$, encoded by the three encoders, respectively. $E_{id}$ is a pre-trained face recognition model without further training so $f_{id}$ owns no expression or pose information. $E_{pose}$ and $E_{exp}$ are trained with corresponding labels, followed by classifiers $C_{p}$ and $C_{exp}$. Furthermore, we adopt an adversarial loss to ensure the learned expression feature is pose-irrelevant. The disentanglement is firstly achieved in feature level. For the overall feature, the decoder is expected to recover the inputting expressional image. When the expression component is removed, a neutral face of the input should be generated with feature $f_{i-p}$. Thus, the expressional component only remains in the expression branch and is identity-invariant and pose-invariant. In this way, the disentanglement is further achieved in image level. A discriminator $D$ is adopted to evaluate the expression category of synthesized images.
Our main contributions are as follows:
\begin{itemize}
  \item We propose a novel model named IPD-FER to disentangle expression component from both identity and head pose. A pre-trained face recognition model is utilized to serve as extra identity or neutral information. It alleviates the requirement of identity labels or neutral-expressional image pairs in previous identity-disentangling FER works and thus makes the disentanglement practicable on in-the-wild databases. 
  \item The GAN-based IPD-FER is able to visualize how the expression component is disentangled by comparing synthesized images. The disentanglement not only reflects in feature level but also in image level, which makes it more reliable. 
  \item Our model achieves state-of-the-art performance not only on easy and lab-controlled databases, but also on complex and diverse in-the-wild expression data.
\end{itemize}


%
%
%
%
\section{Related Work}
Facial expression recognition is challenging due to the subtle variations between different categories and data uncertainty. A spatial-temporal recurrent neural network\cite{zhang2018spatial} is proposed to leverage both electroencephalogram (EEG) signals and video face signals to boost recognition. FDRL\cite{ruan2021feature} models the facial expression as a combination of expression similarities and expression-specific variations and uses feature decomposition and reconstruction learning method to extract fine-grained expression features. \cite{she2021dive} and \cite{zhang2021relative} consider data uncertainty caused by annotation
ambiguity and suppress it by sample reweighting and pairwise uncertainty estimation. \cite{jiang2021boosting} tackles with label noise by selecting potential reliable samples with a teacher-student model. Besides, expressional images are usually entangled with other facial attributes, such as identity and pose, which limit the discrimination of learned deep feature. In this section, we mainly discuss identity-invariant FER and pose-robust FER.
\subsection{Identity-Invariant FER}
Under the assumption that a facial expression is the combination of a neutral face image and the expressive component, DeRL \cite{yang2018facial} generates the corresponding neutral face image for any input face image and utilizes the expression information retained in the GAN model for classification. Conversely, given a facial image with specific identity and expression, \cite{yang2018identity} firstly train conditional generative models to generate six prototypic facial expressions with the same identity. Then a pre-trained CNN is used to extract the features of both the query image and regenerated images. Decision is made by comparing the distances in the feature sub-space. ADFL \cite{bai2019disentangled} extracts the expressive component and neutral component with two different branches and then the difference between the two groups of feature maps are used for classification. \cite{huang2021identity} firstly uses StarGAN to synthesize complete basic emotional facial images for each identity for augmentation, and then metric learning is incorporated for learning more discriminative features. However, the above-mentioned methods need databases that different expressions of one individual are provided, which is hard to meet for in-the-wild data. 

TDGAN \cite{xie2020facial} proposes a expression transferring framework which can generate an expression image to a given face (identity). This is achieved with two branches of encoders, one decoder and two discriminators. The decoder generates a image with the features from two encoders, the discriminators are designed for classifying identities and expressions for the synthesized image respectively. By adversarial learning, the two components are disentangled. Finally only the expression encoder is used for FER task. TER-GAN \cite{ali2021facial} also plays the expression transferring game, but it adds two discriminators for features extracted from two encoders to further ensure the disentanglement. Though these methods doesn't require a specific database that contains different expressions of different individuals, they perform badly on in-the-wild datasets. This is mainly due to the high diversity of spontaneous expression and the effect of other irrelevant factors such as head pose.

Recently, an identity-free conditional Generative Adversarial Network \cite{cai2021identity} was designed to transfer a given facial image to an "average" identity face with the same expression. At the same time, a classifier trained with real images, average faces, and synthesized average images are used for predicting the expression. DLN \cite{zhang2021learning} regards the expression representation as the deviation of overall facial feature and identity feature. A pre-trained face model is utilized to extract identity feature. However, the derived feature is still entangled with head pose. In this paper, we aim to disentangle expression from both identity and pose and boost the recognition.

\subsection{Pose-Robust FER}
To alleviate the influence of variations in head pose, \cite{zhang2018joint} trains the facial expression recognition model with multiple poses to make it more robust. It firstly disentangles identity from expression and pose, then automatically generates face images with different expressions under arbitrary poses. Those generated images are utilized to enrich the training set for FER. A specific database which contains images with various head pose of one person is necessary. 
A discriminative shared Gaussian process latent variable model (DS-GPLVM) is proposed in \cite{eleftheriadis2014discriminative} for multi-view and view-invariant classification of facial expressions from multiple views. DS-GPLVM learns a discriminative manifold shared by multiple views of a facial expression firstly and then conducts facial expression classification in the expression manifold. 
In \cite{zhang2016deep}, the scale invariant feature transform (SIFT) features are extracted corresponding to a set of landmark points of each facial image. The extracted features are then inputted to a well-designed deep neural network to learn discriminative high-level features for expression classification.
A locality-constrained linear coding based bi-layer (LLCBL) model is proposed in \cite{wu2017locality}. In the first layer of model, they estimate head pose and obtain corresponding features from inputting images. Then the features are delivered to a view-dependent model (the second layer) according to the estimated pose for further feature extraction. The features of two layers are combined for training the expression recognition classifier.

\cite{wang2019identity} is the first work which takes both identity and pose into account jointly. It represents the facial feature with one encoder and uses three discriminators to classify identity, pose and expression, respectively. A adversarial learning strategy is adopted that the learned feature should fool the identity and pose discriminator so that it's only expression-related. Besides, a generator is used to reconstruct facial images and thus further favoring the feature representations. However, the identity label is not available for in-the-wild expression data, which limits the capacity of the model. We also aim to learn identity- and pose-invariant representation for FER. Differently, the identity labels are not needed. Our method takes full advantage of the pre-trained model for face recognition task to encode identity information. Expression and pose information are represented by other two independent encoders. The three branches work cooperatively to encode input faces and only the expression branch is utilized for recognition.


\section{Method} 
\begin{figure*}[http]
  \centering
  \includegraphics[scale=0.35]{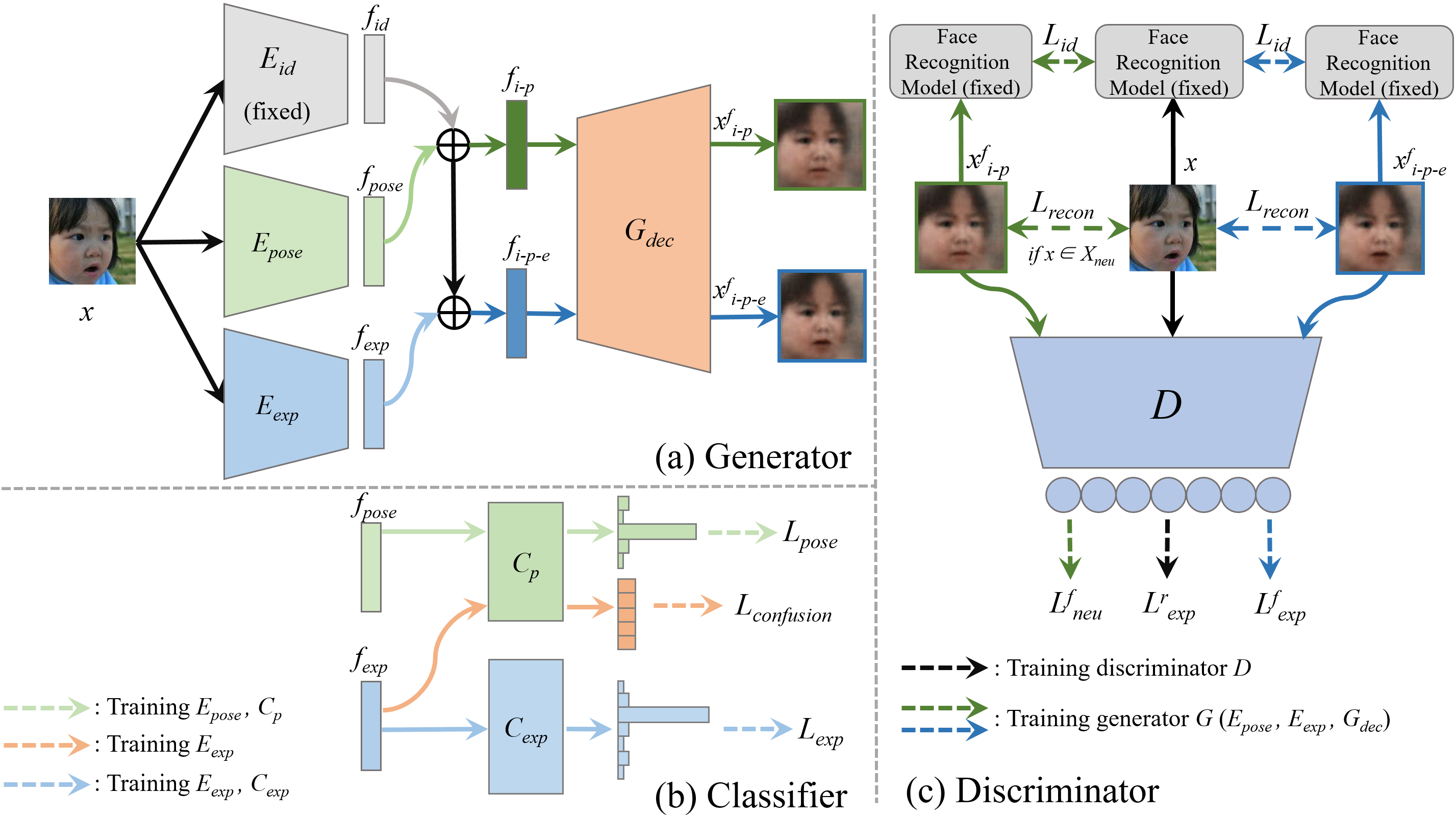}
  \caption{The framework of IPD-FER. The basic idea is to encode the holistic facial feature as the sum of identity, pose, and expression component. (a) The generator consists of three different encoders and one shared decoder. Identity encoder $E_{id}$ is a pre-trained face recognition model with fixed weights, which encodes the input face to an identity feature unrelated to expression and head pose. Pose encoder $E_{pose}$ and expression encoder $E_{exp}$ are trained on expression datasets with corresponding label information, followed by pose classifier $C_{p}$ and expression classifier $C_{exp}$, respectively. Combining both identity feature $f_{id}$ and pose feature $f_{pose}$, the generator should generate an image with the same identity and head pose as input, but with neutral expression. When combining three components $f_{id}$, $f_{pose}$, and expression feature $f_{exp}$, an expressional image almost same as input should be generated. By comparing the two synthesized images, the expression component only remains in the expressional branch and is disentangled from identity and pose. (b) Training with cross-entropy loss, classifiers $C_{p}$ and $C_{exp}$ help encoders to learn pose and expression feature. To further ensure learned expression embedding is pose-invariant, adversarial learning is adopted that the expression encoder should confuse the pose classifier to output uniform predictions. (c) The discriminator $D$ is trained with real images $x$ to predict the expression label. Two synthesized fake images should fool $D$ to classify them into neutral and corresponding expression, respectively. $L_{recon}$ represents reconstruction loss, which facilitates synthesizing images. A pre-trained face recognition model is used to extract identity feature for identity consistency measurements, so as to further ensure the generated images belong to the same individual. }
  \label{method}
  \end{figure*}
We give a detailed introduction of our proposed IPD-FER in this section. We aim to disentangle expression from other adverse factors, i.e., identity and pose. The three attributions are encoded independently and finally only the expression branch is used for recognition. As illustrated in Fig.\ref{method}, the framework consists of one generator $G$, one discriminator $D$, two classifiers $C_{exp}$ and $C_{p}$. Given a facial expressional image $x$ with expression label $y_e$ and pose label $y_p$, we want the generator $G$ to synthesize its fake expressional version $x_{i-p-e}^{f}$ and corresponding neutral version $x_{i-p}^{f}$. So the difference between them is only relevant to expression component and represents more discriminative feature for FER.

\subsection{Generator}
Concretely, the generator owns one decoder $G_{dec}$ and three encoders, i.e., identity encoder $E_{id}$, pose encoder $E_{pose}$ and expression encoder $E_{exp}$. $E_{id}$ derives from a pre-trained face recognition network and is fixed during our training procedure. The well-trained identity encoder disentangles identity from both expression and pose information, with the help of millions of training data. Therefore, feature $f_{id}$ extracted from it is expected to be only relevant to identity, and thus we can consider $f_{id}$ as a frontal and neutral expression representation of the input. The pose encoder and expression encoder are trained with pose labels and expression labels, respectively. By combining $f_{id}$ and $f_{pose}$, $f_{i-p}$ contains both identity and pose information, and should generate the corresponding neutral face $x_{i-p}^{f}$ by the decoder. Similarly, by adding the expression component $f_{exp}$, the decoder should reconstruct the input $x_{i-p-e}^{f}$ from $f_{i-p-e}$. 

To generate desired image pairs, we apply reconstruction loss to make sure that the synthesized images are similar to input in appearance. $\mathcal{L}_{1}$ loss is used to measure the similarity of images. The reconstruction loss is defined as 
\begin{equation}
  \begin{aligned}
    \mathcal{L}_{recon} = \mathbb{E}_{(x,y_e)} \lVert x_{i-p-e}^{f}-x \rVert_1 + \mathbb{E}_{(x,y_e)} \mathbbm{1}[y_e=c] \lVert x_{i-p}^{f}-x \rVert_1
  \end{aligned}
\end{equation}
in which $c$ represents the neutral category. $\mathbbm{1}[\cdot]$ is the indicator function.
In the meanwhile, an identity loss is adopted to further ensure the synthesized images are with the same identity as input. We utilize a pre-trained CNN network $N$ for face recognition task to extract features from both real input image and synthesized fake images. The distances between extracted features are expected to be zero nearly. Therefore, the identity consistency loss can be represented as 
\begin{equation}
  \begin{aligned}
    \mathcal{L}_{id} = \mathbb{E}_x \lVert N(x_{i-p-e}^{f})-N(x) \rVert_1 + \mathbb{E}_x \lVert N(x_{i-p}^{f})-N(x) \rVert_1
  \end{aligned}
\end{equation}

To further separate the identity and expression in latent feature space, we introduce a similarity loss to make them orthogonal to each other. The similarity is measured by cosine distance and we expect the value to be as close as zero. The loss function is denoted as:
\begin{equation}
  \begin{aligned}
    \mathcal{L}_{cos} = \mathbb{E}_{x}\frac{| f_{id}\cdot f_{exp}|}{\vert\vert f_{id} \vert\vert_2 \cdot \vert\vert f_{exp} \vert\vert_2}
  \end{aligned}
\end{equation}

For further ensuring the learned expression feature is pose-irrelevant, we introduce an adversarial loss $\mathcal{L}_{confusion}$ that $f_{exp}$ should fool $C_{p}$ to predict the probability distribution as close as possible to a uniform distribution over the category. The formula is written as
\begin{equation}
  \mathcal{L}_{confusion} = - \mathbb{E}_{x} \sum_{k=1}^{K_{P}} \frac{1}{K_{P}} \log \frac{e^{{C_{p}(E_{exp}(x))}_{k}}}{\sum_{j=1}^{K_{P}}e^{{C_{p}(E_{exp}(x))}_{j}}} 
\end{equation}

\subsection{Classifier}
The identity encoder is fixed during training due to the existence of pre-training, but the other two encoders require supervision to function normally. We use fully connected layer as classifier $C_{exp}$ and $C_{p}$. Cross entropy loss is adopted for them. Encoders and corresponding classifiers are optimized jointly.
\begin{equation}
  \mathcal{L}_{exp} = -\mathbb{E}_{(x,y_e)}\sum_{k=1}^K \mathbbm{1}[y_e=k]\log \frac{e^{{C_{exp}(E_{exp}(x))}_k}}{\sum_{j=1}^{K}e^{{C_{exp}(E_{exp}(x))}_{j}}}
\end{equation}
\begin{equation}
  \mathcal{L}_{pose} = -\mathbb{E}_{(x,y_{p})}\sum_{k=1}^{K_P} \mathbbm{1}[y_{p}=k]\log \frac{e^{{C_{p}(E_{pose}(x))}_k}}{\sum_{j=1}^{K}e^{{C_{p}(E_{pose}(x))}_{j}}}
\end{equation}
$K$ and $K_P$ are the number of expression categories and pose categories, respectively.
We define classification loss $\mathcal{L}_{c}$ as the sum of above two items that \begin{equation}
  \mathcal{L}_{c} = \mathcal{L}_{exp}+\mathcal{L}_{pose}
\end{equation}
Since our aim is recognizing facial expression, the expression encoder $E_{exp}$ and classifier $C_{exp}$ are finally utilized for inference.

\subsection{Discriminator}
The discriminator $D$ is trained to evaluate whether the generated images are with desired expression. It should conduct K-expression recognition. The cross entropy loss computed by the input real images $x$ is used to supervise its training, which is defined as 
\begin{equation}
  \mathcal{L}_{exp}^{r} = -\mathbb{E}_{(x,y_e)}\sum_{k=1}^K \mathbbm{1}[y_e=k]\log(D(x))
\end{equation}
For convenience, $D(\cdot)$ denotes the predicted class distribution after softmax operation.
$G$ should fool $D$ to predict synthesized images as neutral and corresponding expression category. To this end, an adversarial loss is applied. For $G$, the adversarial loss for fake neutral face is written as
\begin{equation}
  \begin{aligned}
    \mathcal{L}_{neu}^{f} = -\mathbb{E}_{(x,c)}\sum_{k=1}^K \mathbbm{1}[k=c]\log(D(G_{dec}(E_{id}(x)\\+E_{pose}(x))))
  \end{aligned}
\end{equation}
where c denotes the neutral ground truth label. The adversarial loss for fake expressional face is written as
\begin{equation}
  \begin{aligned}
    \mathcal{L}_{exp}^{f} = -\mathbb{E}_{(x,y_e)}\sum_{k=1}^K \mathbbm{1}[y_e=k]\log(D(G_{dec}(E_{id}(x)\\+E_{pose}(x)+E_{exp}(x))))
  \end{aligned}
\end{equation}

\subsection{Overall learning}
So far, the expression component is disentangled from identity and pose. Firstly, they are represented with different encoders in the feature level. Secondly, the expression component reflects in the discrepancy between generated neutral and expressional images.
We define the generative loss for $G$ as
\begin{equation}
  \begin{aligned}
    \mathcal{L}_{G}^{'} &= \lambda_{1}\mathcal{L}_{neu}^{f}+\lambda_{2}\mathcal{L}_{exp}^{f}
    +\lambda_{3}\mathcal{L}_{id}+\lambda_{4}\mathcal{L}_{recon}
  \end{aligned}
\end{equation}
and the overall loss for $G$ is 
\begin{equation}
\begin{aligned}
\mathcal{L}_{G} = \mathcal{L}_{G}^{'} + \mathcal{L}_{c} +\beta_{1}\mathcal{L}_{cos}+\beta_{2}\mathcal{L}_{confusion}
\end{aligned}
\end{equation}
$G$ and $D$ are trained in an alternative manner and the learning algorithm is summarized in Algorithm \ref{alg::1}.
\begin{algorithm}[h] 
  \caption{The learning procedure of our method.} 
  \label{alg::1} 
  \begin{algorithmic}[1] 
  \Require 
  Training set in which each image $x$ is labeled with label $\{y,y_{p}\}$ ;
  
  Generator $G$=\{$E_{id}$,$E_{pose}$,$E_{exp}$,$G_{dec}$\}; 
  
  Discriminator $D$; 
  
  Classifiers $C_{exp}$ and $C_{p}$
  
  \Ensure 
  optimal $E_{pose} \circ C_{exp}$ 
  \Repeat 
  \State Fix $D$ and $G_{dec}$, update $E_{pose}$, $C_{p}$, $E_{exp}$ and $C_{exp}$ jointly by gradient descent: 
  
  $\nabla \theta_{E_{pose}\cup C_{p}\cup E_{exp} \cup C_{exp}} := \frac {\partial(\mathcal{L}_{c}+\beta_1\mathcal{L}_{cos})}{\partial\theta_{E_{pose}\cup C_{p}\cup E_{exp} \cup C_{exp}}}$; 
  \State Fix $G$, $C_{exp}$ and $C_{p}$, update $D$ by gradient descent: $\nabla \theta_{D} := \frac {\partial\mathcal{L}_{exp}^{r}}{\partial\theta_{D}}$; 
  \State Fix $D$, $C_{exp}$ and $C_{p}$, update $G$ by gradient descent: $\nabla \theta_{G} := \frac {\partial(\mathcal{L}_{G}^{'}+\beta_2\mathcal{L}_{confusion})}{\partial\theta_{G}}$; 
  \Until {Converge}
  \end{algorithmic} 
\end{algorithm}

\section{Experiments}
In this section, we evaluate the effectiveness of our method on one lab-controlled and four in-the-wild databases. They are CK+, RAF-DB, FERPlus, AffectNet, and SFEW 2.0, respectively.

\subsection{Databases}
\textbf{CK+} \cite{lucey2010extended}: The lab-controlled database Extended CohnKanade (CK+) contains 593 video sequences from 123 subjects. These videos are labeled to 7 classes (anger, contempt, disgust, fear, happiness, sadness, and surprise). We extract the final three frames of each sequence with peak formation, resulting in 981 images. In order to obtain identity information, 327 neutral images collected from the first frame of each sequence are also used. Then we divided the selected images into 10 folds by ID to conduct person-independent 10-fold cross-validation.

\textbf{RAF-DB} \cite{li2018reliable}: It is a large-scale in-the-wild facial expression database which contains about 30,000 great diverse facial images from thousands of individuals downloaded from the Internet. Images in RAF-DB were labeled by 315 human coders and the final annotations were determined through the crowdsourcing techniques. And each image was assured to be labeled by about 40 independent labelers. RAF-DB contains 12,271 training samples and 3,068 test samples annotated with seven basic emotional categories (i.e., angry, disgust, fear, happy, neutral, sad and surprise). Compound expressions labeled as 11 classes are also provided, which are not used in our experiment.

\textbf{FERPlus} \cite{barsoum2016training}: The FERPlus database is an extension of FER2013 \cite{goodfellow2013challenges}. The large-scale and unconstrained database FER2013 was created and labeled automatically by the Google image search API. All images in FER2013 have been registered and resized to 48$\times$48 pixels. FER2013 contains 28,709 training images, 3,589 validation images and 3,589 test images with seven expression labels. It’s relabeled in 2016 by Microsoft with each image labeled by 10 individuals to consist 8 classes (contempt is added), thus has more reliable annotations. 

\textbf{AffectNet} \cite{mollahosseini2017affectnet}: The large-scale AffectNet database contains more than 0.4 millions of labeled images. The images are downloaded from Internet using three search engines and expression-related keywords. It’s the largest dataset for FER currently. Similar to RAF-DB, we use the labeled images in seven basic expression categories, resulting in about 280,000 training samples. There are 500 samples per category in validation set, resulting 3500 validation images. 

\textbf{SFEW 2.0}\cite{dhall2015video}: The Static Facial Expressions in the Wild (SFEW) \cite{dhall2011static} was created by selecting static frames from the AFEW database by computing key frames based on facial point clustering. The data are collected from movie clips, with various head poses, occlusions and illuminations. SFEW 2.0 is the most commonly used version of it, which is the benchmark for the SReco sub-challenge in EmotiW 2015. It is divided into three sets: Train (958 samples), Val (436 samples) and Test (372 samples). Each image is annotated to one of seven basic expressions, as that in RAF-DB. Since the labels of test set are not available, we use the validation set for testing.

\subsection{Implementation Details}

\begin{figure}[]
\centering
\includegraphics[scale=0.35]{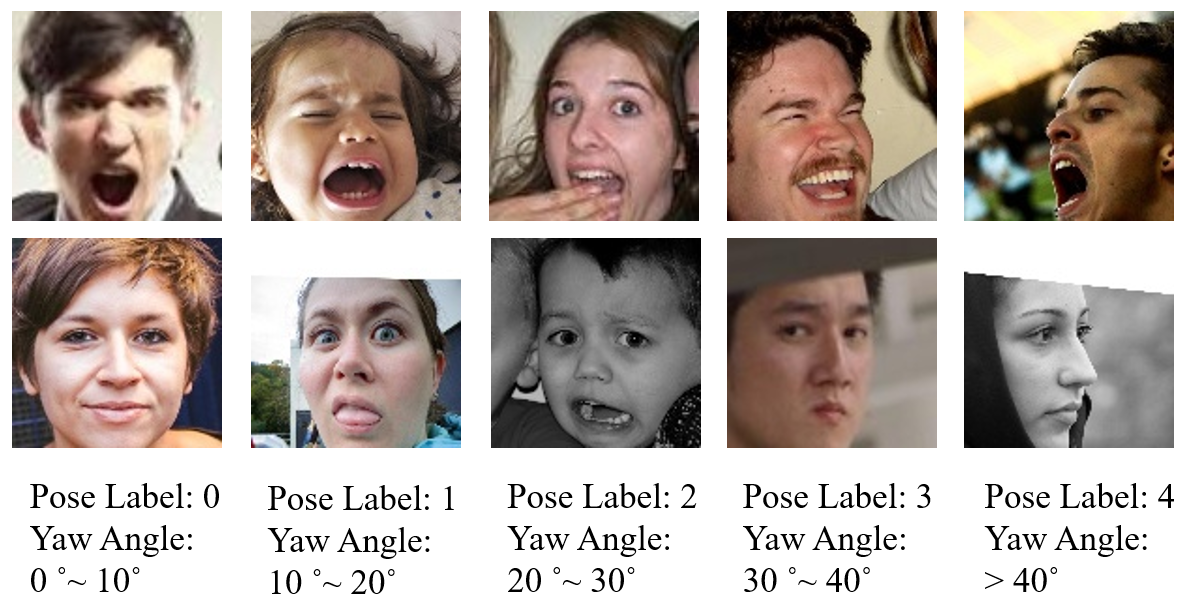}
\caption{Demonstration of images with corresponding pose label in RAF-DB.}
\label{vis_pose}
\end{figure}

To make full use of pre-trained face recognition network, the three encoders and discriminator $D$ derive from ResNet-18 which is pre-trained on CASIA-WebFace\cite{yi2014learning} database. The weights of identity encoder are fixed during training. The dimension of their outputting features is 512. $C_{exp}$ and $C_{p}$ are fully connected layers. $G_{dec}$ consists of 5 layers of sequence of transpose convolution, Instance Normalization, ReLU and ResNet Block. It converts the 512-dimensional feature vector to a 3$\times$112$\times$112 image. All networks are implemented by Pytorch. For each image in all databases, the facial region is firstly detected by MTCNN \cite{zhang2016joint} and then resized and normalized to 112$\times$112 pixels with similarity transform. 
To obtain pose labels for in-the-wild databases, a trained head pose estimator is utilized.\footnote{https://github.com/WIKI2020/FacePose\_pytorch} Five pose categories are obtained for these databases, as shown in Table \ref{label_pose}. Some examples are shown in Fig. \ref{vis_pose}. And we regard images whose yaw angle between $0^\circ$ and $10^\circ$ as frontal faces.
Since images in CK+ are all frontal faces, we remove the pose encoder and corresponding discriminator and only encode the facial representation as the sum of identity and expression.
We set the batch size to 32 and use Adam optimizer. The learning rate begins with 0.0001 and decreases by 0.1 every 10 epochs. Training lasts for 30 epochs in total. The hyper-parameters $\lambda_{1}$, $\lambda_{2}$, $\lambda_{3}$, $\lambda_{4}$, $\beta_{1}$, $\beta_{2}$ are set to 0.001, 0.001, 1, 10, 0.5, 1, respectively. For RAF-DB, FERPlus, AffectNet and SFEW 2.0, we repeat the experiments for 4 times with different random seeds, and report the best and the average validation performance.
\begin{table}[]
  \centering
  \caption{Head pose information of training set of RAF-DB, FERPlus, AffectNet, and SFEW 2.0.}
  \label{label_pose}
  \renewcommand\tabcolsep{2.0pt} 
  \begin{tabular}{cccccc}
  \toprule
  Pose Label & Yaw Angle  & RAF-DB  & FERPlus & AffectNet & SFEW 2.0 \\
  \midrule
  0 & $0^\circ\sim10^\circ$ & 5836 & 13515 & 167448 & 453\\
  1 & $10^\circ\sim20^\circ$ & 2544 & 4985 & 66322 & 279\\
  2 & $20^\circ\sim30^\circ$ & 1646 & 3200 & 35732 & 157\\
  3 & $30^\circ\sim40^\circ$ & 1058 & 2486 & 12146 & 49\\
  4 & \textgreater$40^\circ$ & 1187 & 874  & 2253 & 17\\
  \bottomrule
  \end{tabular}
  \end{table}

\subsection{Comparison with state-of-the-arts}
\begin{figure}[h]
    \centering
    \subfigure[6 classes]{\includegraphics[scale=0.25]{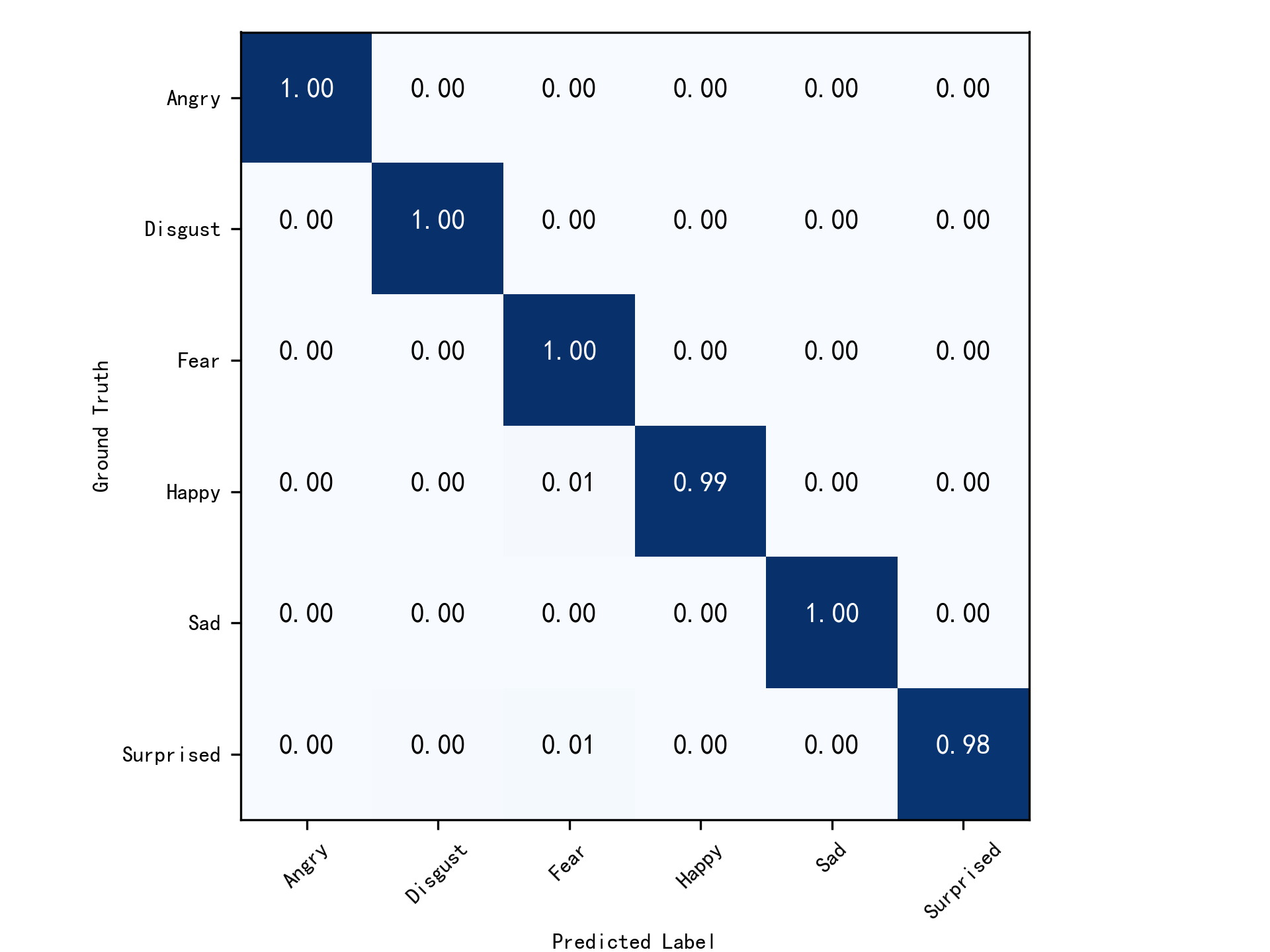}}
    \subfigure[7 classes]{\includegraphics[scale=0.25]{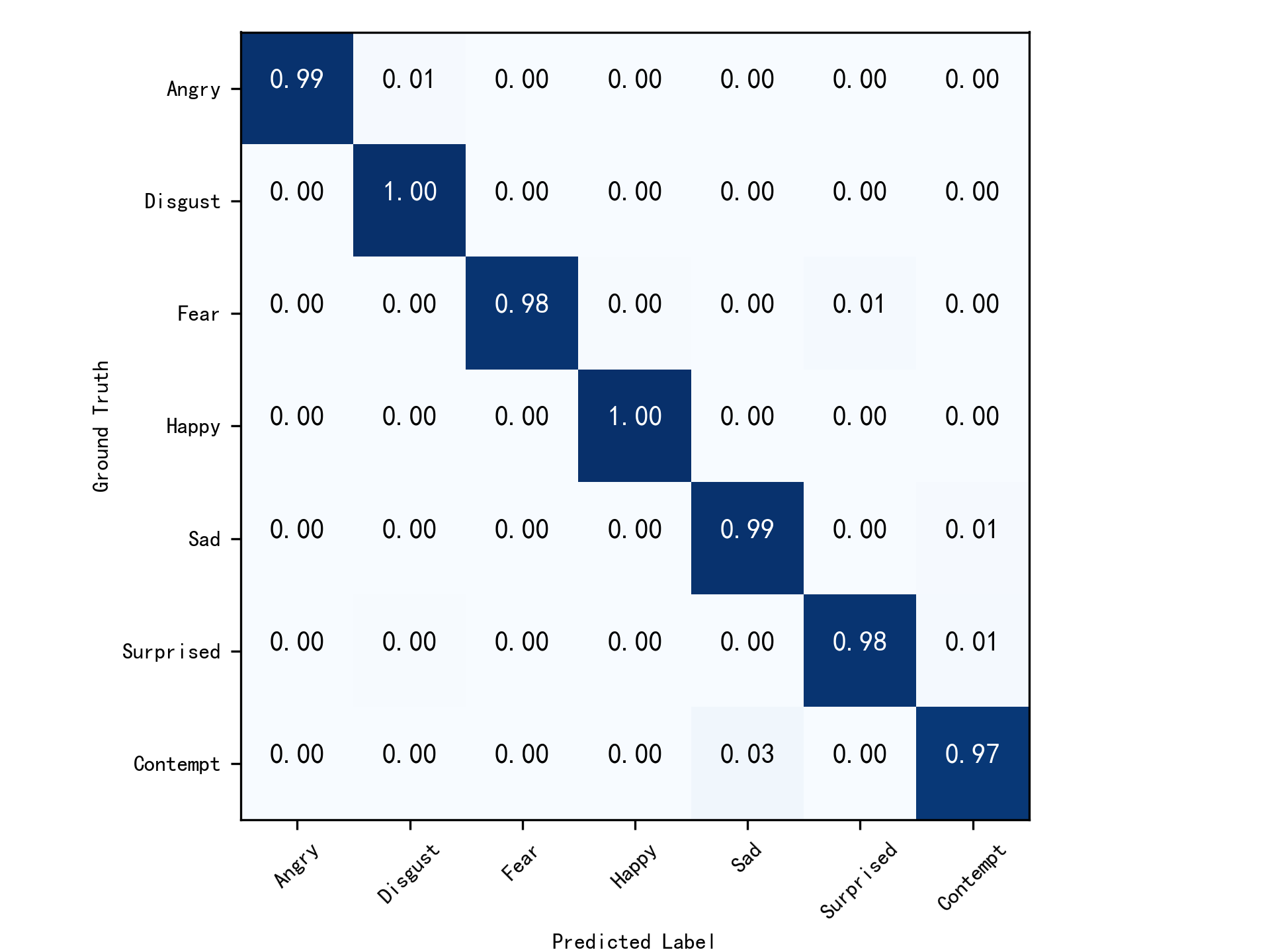}}
    \caption{The confusion matrices on CK+ by IPD-FER. }
    \label{cm_CK}
  \end{figure}

In this section, we compare our IPD-FER with other state-of-the-art (SOTA) methods. Since almost existing works study how to alleviate the effect of identity conduct experiments only on lab-controlled databases. We compare those methods with IPD-FER on CK+ in Table \ref{res_CK}. The confusion matrices are shown in Fig. \ref{cm_CK}. Our disentangling strategy outperforms previous works. The baseline for which we fine-tune the pre-trained ResNet-18 on CK+ is also provided. Although simple fine-tuning strategy makes considerable result, our method also outperforms it because the disentanglement of identity and expression enables the expression encoder to extract more discriminative features. 
\begin{table}[h]
  \centering
  \caption{The comparison of accuracies on CK+ database.}
  \label{res_CK}
  \begin{tabular}{ccc}
  \toprule
  Methods   & Setting                    & Accuracy (\%)    \\
  \midrule
  IDFERM \cite{liu2019hard}    & 6 expressions and contempt & 98.35            \\
  DeRL \cite{yang2018facial}      & 6 expressions and contempt & 97.3             \\
  IA-gen \cite{yang2018identity}    & 6 expressions and contempt & 96.57            \\
  ADFL \cite{bai2019disentangled}      & 6 expressions and contempt & 98.17            \\
  TDGAN \cite{xie2020facial}     & 6 expressions and neutral  & 97.53            \\
  TER-GAN \cite{ali2021facial}   & 6 expressions and contempt & 98.47            \\
  Huang et.al. \cite{huang2021identity}  & 6 expressions and contempt & 98.65            \\
  IF-GAN \cite{cai2021identity}   & 6 expressions              & 97.52            \\
  Cross-VAE \cite{wu2020cross} & 6 expressions and contempt & 94.96            \\
  FN2EN \cite{ding2017facenet2expnet}     & 6 expressions              & 98.6             \\
  \hline
  ResNet-18\dag & 6 expressions              & 98.22            \\
  ResNet-18\ddag & 6 expressions and contempt & 97.39            \\
  IPD-FER\dag  & 6 expressions              & \textbf{99.28}   \\
  IPD-FER\ddag  & 6 expressions and contempt & \textbf{98.65}   \\
  \bottomrule
  \end{tabular}
  \end{table}
  
\begin{table}[h]
\centering
\caption{The comparison of accuracies on RAF-DB database.}
\label{res_RAF}
\begin{tabular}{ccc}
\toprule
Methods     & Accuracy (\%)    \\
\midrule
TDGAN \cite{xie2020facial}      & 81.91            \\
IF-GAN \cite{cai2021identity}   & 88.33            \\
DLN \cite{zhang2021learning}    & 86.4           \\
Cross-VAE \cite{wu2020cross}    & 84.81            \\
IPA2LT(LTNet)\cite{zeng2018facial} & 86.77        \\
RAN-ResNet18 \cite{wang2020region}  & 86.90       \\
SCN\cite{wang2020suppressing} & 88.14  \\
Progressive Teacher\cite{jiang2021boosting} & 88.69  \\
\hline
ResNet-18  & 87.48 (87.32±0.16)           \\
IPD-FER   & \textbf{88.89} (88.67±0.15)   \\
\bottomrule
\end{tabular}
\end{table}
  
\begin{table}[h]
\centering
\caption{The comparison of accuracies on FERPlus database.}
\label{res_FER}
\begin{tabular}{ccc}
\toprule
Methods     & Accuracy (\%)    \\
\midrule
PLD \cite{barsoum2016training}      & 85.1            \\
ResNet+VGG \cite{huang2017combining} & 87.4            \\CNNs and BOVW + local SVM \cite{georgescu2019local} & 87.76         \\
SCN \cite{wang2020suppressing}   & 88.01         \\
\hline
ResNet-18  & 87.75 (87.63±0.11)        \\
IPD-FER   & \textbf{88.42} (88.32±0.09) \\
\bottomrule
\end{tabular}
\end{table}

\begin{table}[h]
\centering
\caption{The comparison of accuracies on AffectNet database.}
\label{res_AffectNet}
\begin{tabular}{ccc}
\toprule
Methods     & Accuracy (\%)    \\
\midrule
IPA2LT \cite{zeng2018facial}   &   56.51       \\
gACNN\cite{li2018occlusion}   &    58.78     \\
IPFR\cite{wang2019identity}      &      57.4       \\
RAN-ResNet18 \cite{wang2020region}     &   59.5       \\
Chen et.al. \cite{chen2021residual} & 61.98\\
\hline
ResNet-18  & 59.06 (58.74±0.22)      \\
IPD-FER    & \textbf{62.23} (61.98±0.19)     \\
\bottomrule
\end{tabular}
\end{table}
  
\begin{table}[h]
\centering
\caption{The comparison of accuracies on SFEW 2.0 database.}
\label{res_SFEW}
\begin{tabular}{ccc}
\toprule
Methods     & Accuracy (\%)    \\
\midrule
IACNN \cite{meng2017identity}      & 50.98            \\
FN2EN \cite{ding2017facenet2expnet}   & 55.15         \\
IL-CNN \cite{cai2018island}      & 52.52            \\
RAN-ResNet18 \cite{wang2020region}  & 54.19            \\
Covariance Pooling \cite{acharya2018covariance}  & 58.14            \\
\hline
ResNet-18  & 46.88 (45.79±0.82)      \\
IPD-FER    & 52.43 (50.69±1.35)     \\
\hline
ResNet-18(pre-trained)  & 56.12 (55.25±0.50)    \\
IPD-FER(pre-trained)    & \textbf{58.43} (58.05±0.11)  \\
\bottomrule
\end{tabular}
\end{table}
For RAF-DB, methods \cite{xie2020facial}, \cite{cai2021identity}, \cite{zhang2021learning} and \cite{wu2020cross} all take identity or pose into account, the result is reported in Table \ref{res_RAF}. Although the expression transferring strategy performs considerably on CK+, it doesn't work well on RAF-DB due to the complexity of in-the-wild expression data. DLN regards the expression component as deviation of learned expressional feature and identity feature, but it neglects the effect of head pose. We also outperforms IF-GAN which removing the identity component by transferring the query image to a category-average face. With our proposed IPD-FER, the accuracy on RAF-DB is superior than baseline by 1.41\%. 

\begin{figure*}[http]
  \centering
  \subfigure[RAF-DB]{\includegraphics[scale=0.16]{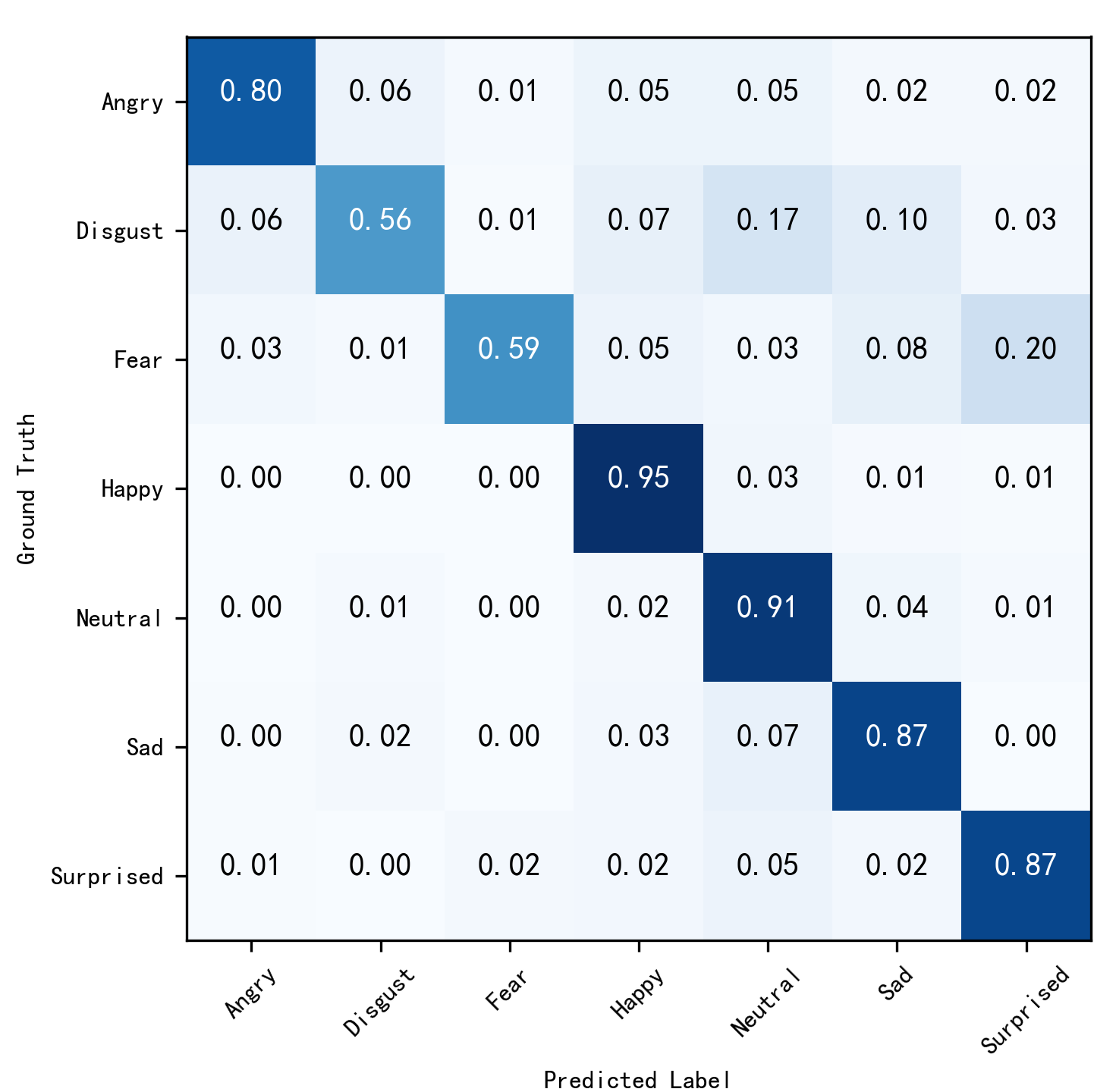}}
  \subfigure[FERPlus]{\includegraphics[scale=0.16]{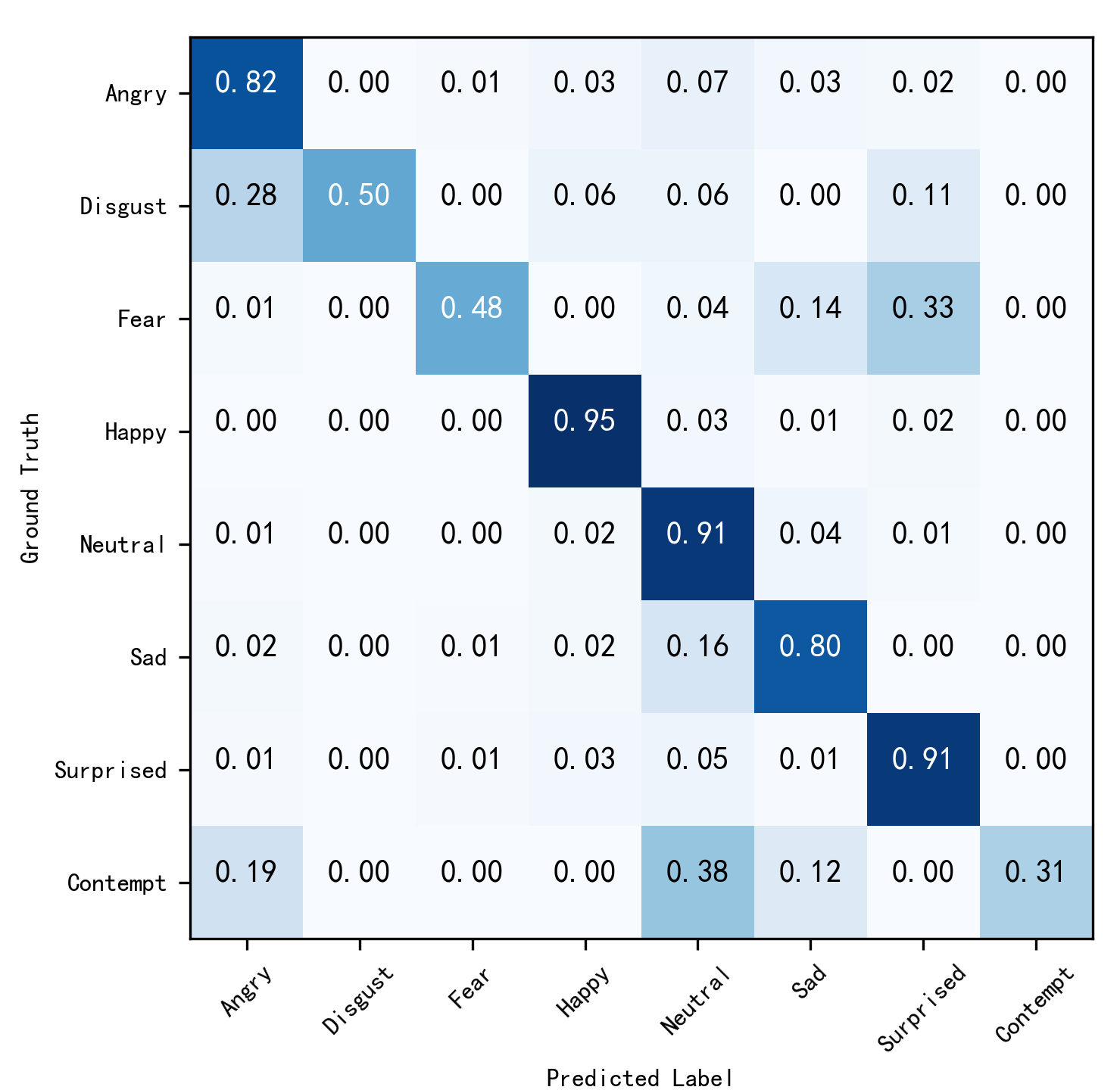}}
  \subfigure[AffectNet]{\includegraphics[scale=0.16]{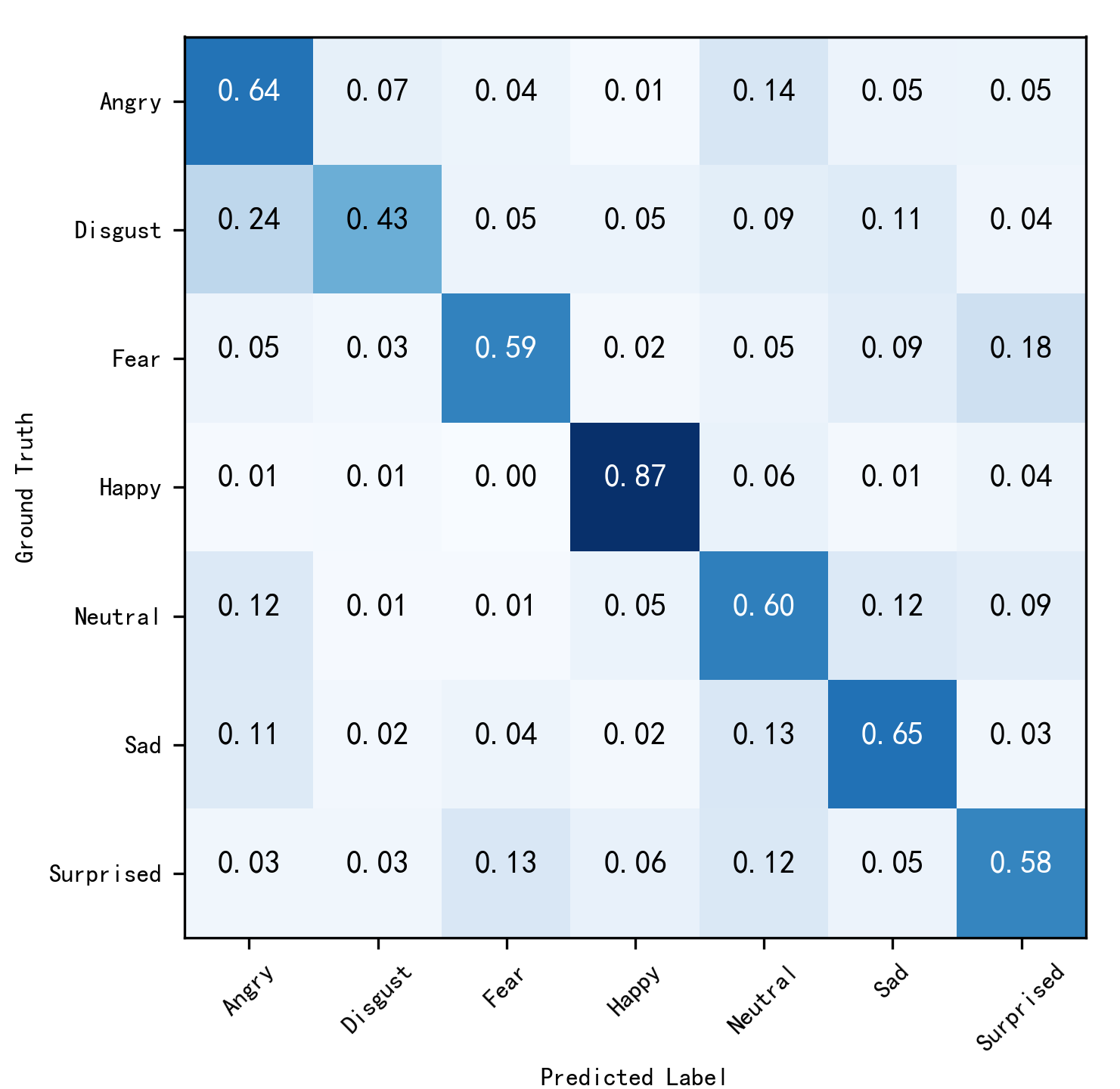}}
  \subfigure[SFEW 2.0]{\includegraphics[scale=0.33]{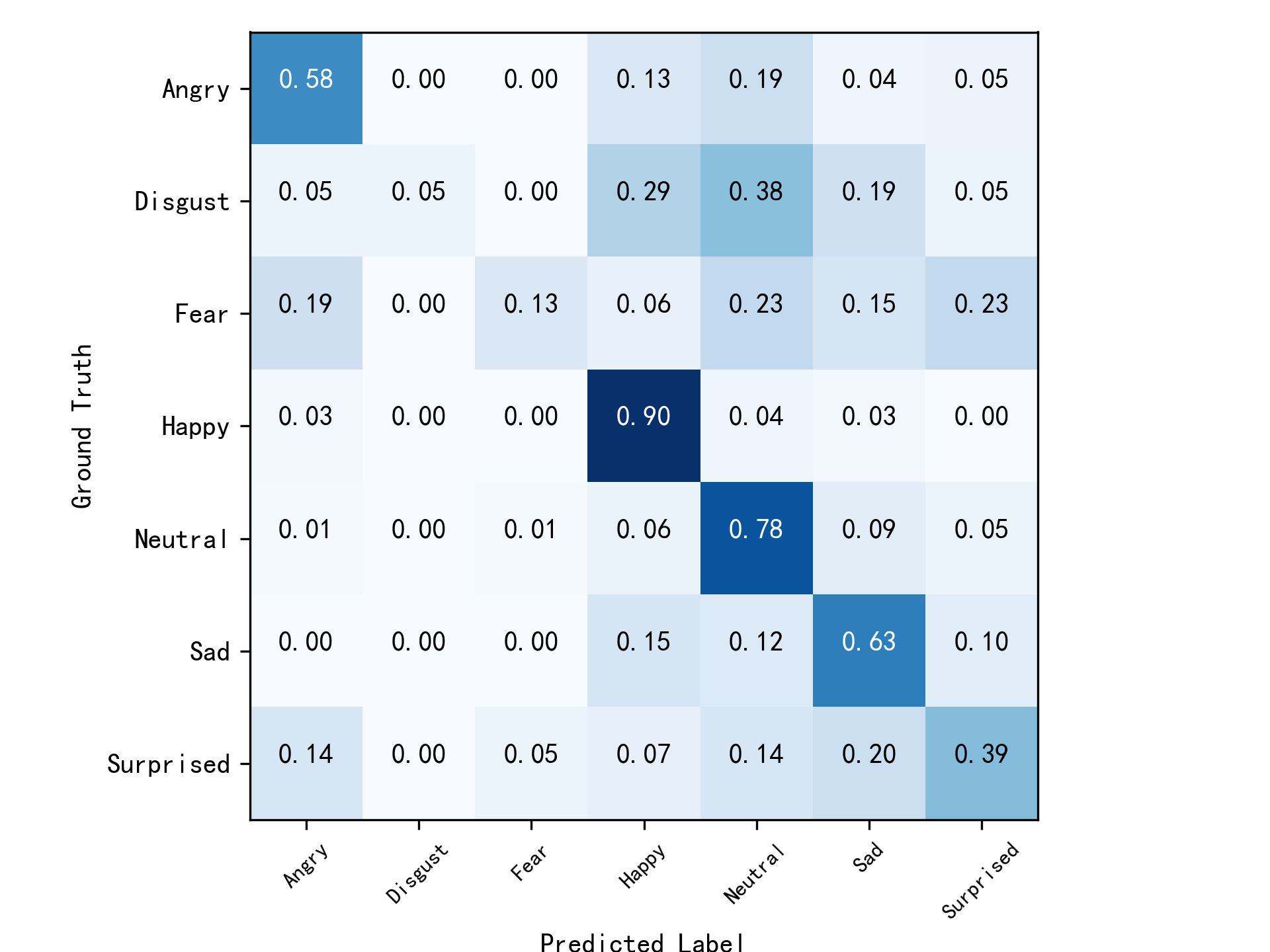}}
  \caption{The confusion matrices on four in-the-wild databases. }
  \label{cm_all}
\end{figure*}

For FERPlus, AffectNet and SFEW2.0, we compare our method with other SOTA approaches in Table \ref{res_FER}, Table \ref{res_AffectNet} and Table \ref{res_SFEW}, respectively. IPD-FER improves baseline by 0.67\% on FERPlus and 3.17\% on AffectNet. Because SFEW 2.0 contains limited training data, we use the pre-trained model on RAF-DB and fine-tune it. The accuracy is improved from 56.12\% to 58.43\%. Without pre-training, the performance of IPD-FER is also superior than baseline by 5.55\% on SFEW 2.0 database. Our approach also achieves better or comparable results than SOTA. We compute the confusion matrices of four in-the-wild databases and show them in Fig. \ref{cm_all}. 

We also compare the efficiency of IPD-FER with baseline (ResNet-18). Table \ref{res_time} shows the consumed time of training. We use one GeForce GTX 1080 GPU for each experiment. Although our method takes longer time for training, it improves the recognition performance on all databases. Additionally, we only use the expression encoder $E_{exp}$ and classifier $C_{exp}$ for final classification. Therefore, it doesn't bring any time cost for inference when compared with baseline. Taking the 7-class classification for example, the two methods both take about 2 ms for each image when using GPU (GeForce GTX 1080) for inference. The inference speed is 454 im/s. When only using CPU (Intel(R) Core(TM) i7-7820X CPU @ 3.60GHz) for inference, it takes about 40 ms for each image, the inference speed is 25 im/s.
\begin{table}[h]
  \centering
  \caption{The comparison of training time on different databases.}
  \label{res_time}
  \begin{tabular}{ccc}
  \toprule
  Database    &  ResNet-18  & IPD-FER \\
  \midrule
  RAF-DB    &  0.5 h  & 2.5 h \\
  FERPlus    &  1 h  & 7.5 h \\
  AffectNet    &  2.5 h  & 24 h \\
  SFEW 2.0    &  0.05 h  & 0.3 h \\
  \bottomrule
  \end{tabular}
\end{table}

\subsection{Ablation Study}
\begin{figure}[h]
  \centering
  \includegraphics[scale=0.6]{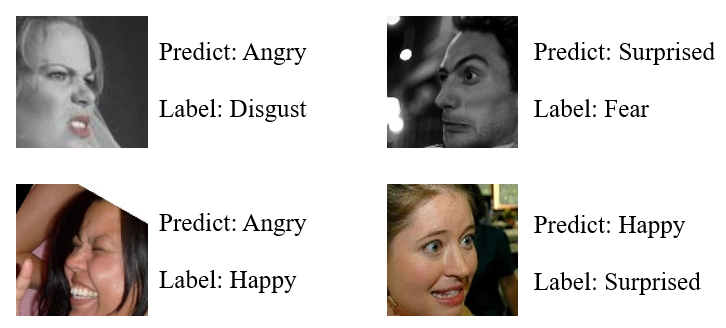}
  \caption{A visualization of some large-pose samples misclassified by ID-FER but correctly predicted by IPD-FER in RAF-DB test set.} 
  \label{wrong_predict}
\end{figure}

Since in-the-wild databases don't contain identity labels, we only evaluate the effectiveness of IPD-FER on alleviating the influence of head pose. We select faces with yaw angle larger than certain value from RAF-DB test set to construct different test subsets. Four subsets are formed in which the yaw angle of each face is larger than $10^\circ$, $20^\circ$, $30^\circ$, $40^\circ$, respectively. Then we evaluate the recognition performances of baseline, ID-FER and IPD-FER on these test subsets. The pose encoder $E_{pose}$, classifier $C_{p}$ and corresponding loss functions are not used in ID-FER, which means that the expression component is still entangled with head pose. The results are reported in Table \ref{acc_pose}. As the head pose gets larger, the recognition accuracy of baseline model gets lower. When head pose reaches larger than $40^\circ$, it only obtains accuracy of 81.49\%. For ID-FER, its performance is near to baseline model since it doesn't take head pose into account. With the help of pose encoder and adversarial learning, IPD-FER keeps high recognition performance on large-pose test subsets. It achieves accuracy of 87.66\% on subset that head pose is larger than $40^\circ$. We list some large-pose samples in RAF-DB test set that are misclassified by ID-FER but are correctly predicted by IPD-FER in Fig. \ref{wrong_predict}.
\begin{table}[h]
  \centering
  \caption{The performances of different methods on different test subsets of RAF-DB.}
  \label{acc_pose}
  \begin{tabular}{ccccc}
  \toprule
  \multirow{2}{*}{Test set} & \multirow{2}{*}{Number} & \multicolumn{3}{c}{Accuracy (\%)} \\\cline{3-5}
           &\rule{0pt}{10pt}& baseline       & ID-FER & IPD-FER \\
  \midrule
  All      & 3068              & 87.06          & 87.65  & \textbf{88.89}   \\
  Pose(\textgreater$10^\circ$)       & 1704              & 86.44          & 86.74  & \textbf{88.67}   \\
  Pose(\textgreater$20^\circ$)       & 990               & 85.56          & 86.46  & \textbf{88.79}   \\
  Pose(\textgreater$30^\circ$)       & 574               & 83.45          & 85.02  & \textbf{87.11}   \\
  Pose(\textgreater$40^\circ$)       & 308               & 81.49 & 84.09  & \textbf{87.66}   \\
  \bottomrule
  \end{tabular}
\end{table}

To evaluate the effect of utilized loss functions or modules, we design an ablation study to investigate $L_{confusion}$, $L_{cos}$, $L_{id}$, $L_{recon}$, and discriminator $D$ on RAF-DB. The experimental results are shown in Table \ref{res_ablation}. Several observations are derived. Firstly, the recognition performance only degrades by 0.79\% without $L_{confusion}$. This can be explained by that pose information is already represented with corresponding encoder and thus it remains rarely in the expression encoder. We use the confusion loss to further reduce the correlation between them. Secondly, the accuracy degrades by 1.67\% when $L_{cos}$ is not used. This indicates that making identity and expression orthogonal in the feature space benefits their disentanglement. Thirdly, $L_{id}$ further ensures the generated images are with the same identity since only using $L_{recon}$ for reconstruction is not reliable enough. Finally, the discriminator is necessary for synthesizing corresponding images. Even though the reconstruction of generated images can be achieved by $L_{recon}$ and $L_{id}$, the discrepancy between them reflects only in pixel level. $D$ helps compare the generated neutral and expressional images with label guidance and thus benefits disentanglement of identity and expression.

\begin{table}[h]
  \centering
  \caption{Evaluation of different loss functions or module on RAF-DB.}
  \label{res_ablation}
  \begin{tabular}{ccccc}
  \toprule
  $L_{confusion}$ & $L_{cos}$ & $L_{id}$ & $D$ & Accuracy(\%)\\
  \midrule
  \xmark & \checkmark & \checkmark  & \checkmark & 88.20 \\
  \checkmark & \xmark & \checkmark  & \checkmark & 87.32 \\
  \checkmark & \checkmark & \xmark  & \checkmark & 87.58 \\
  \checkmark & \checkmark & \checkmark  & \xmark &  87.81\\
  \checkmark & \checkmark & \checkmark  & \checkmark &  88.89\\
  \bottomrule
  \end{tabular}
\end{table}

\begin{figure}[h]
  \centering
  \subfigure[$\lambda_{1}$ and $\lambda_{2}$]{\includegraphics[scale=0.25]{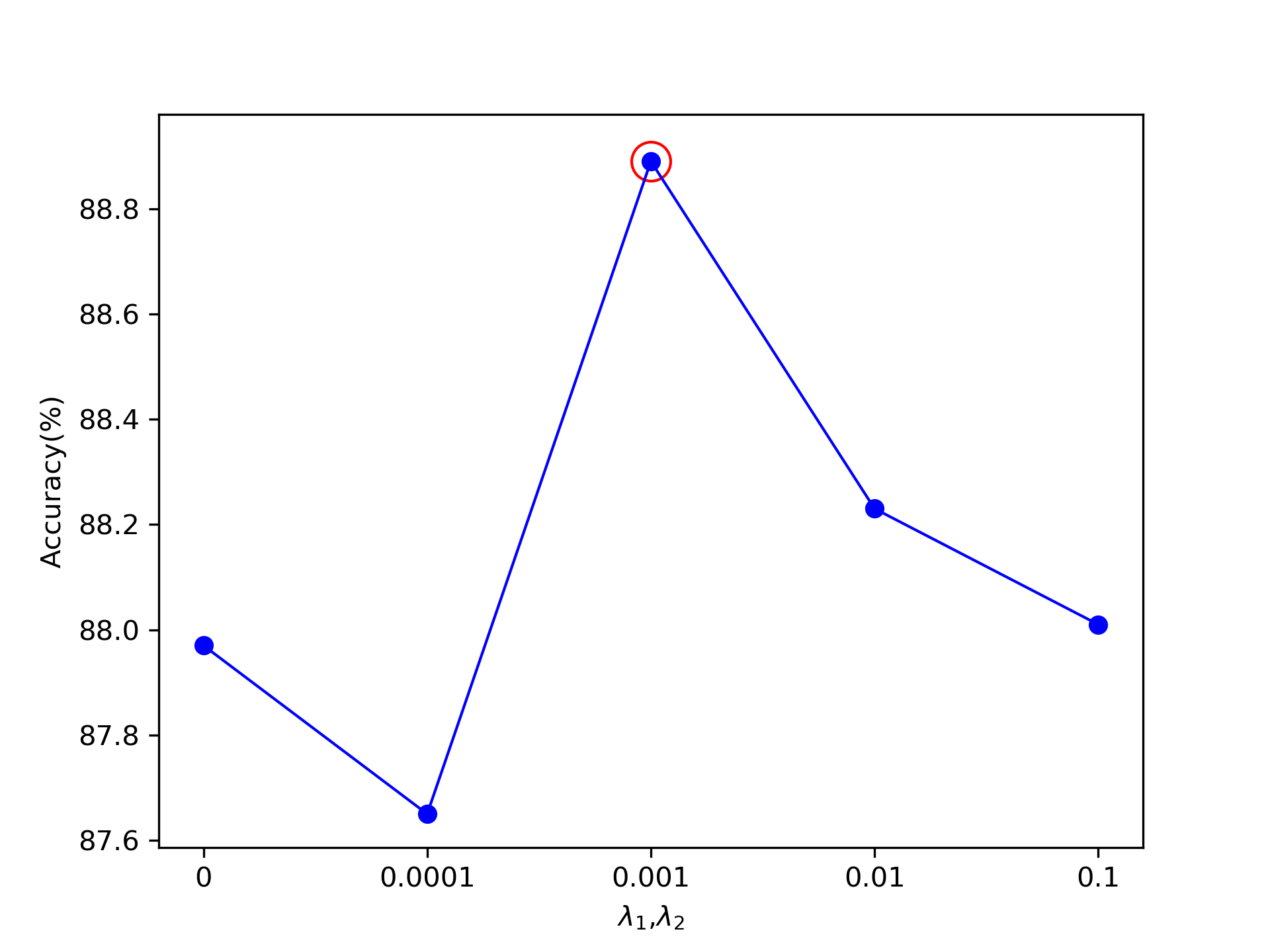}}
  \subfigure[$\lambda_{3}$]{\includegraphics[scale=0.25]{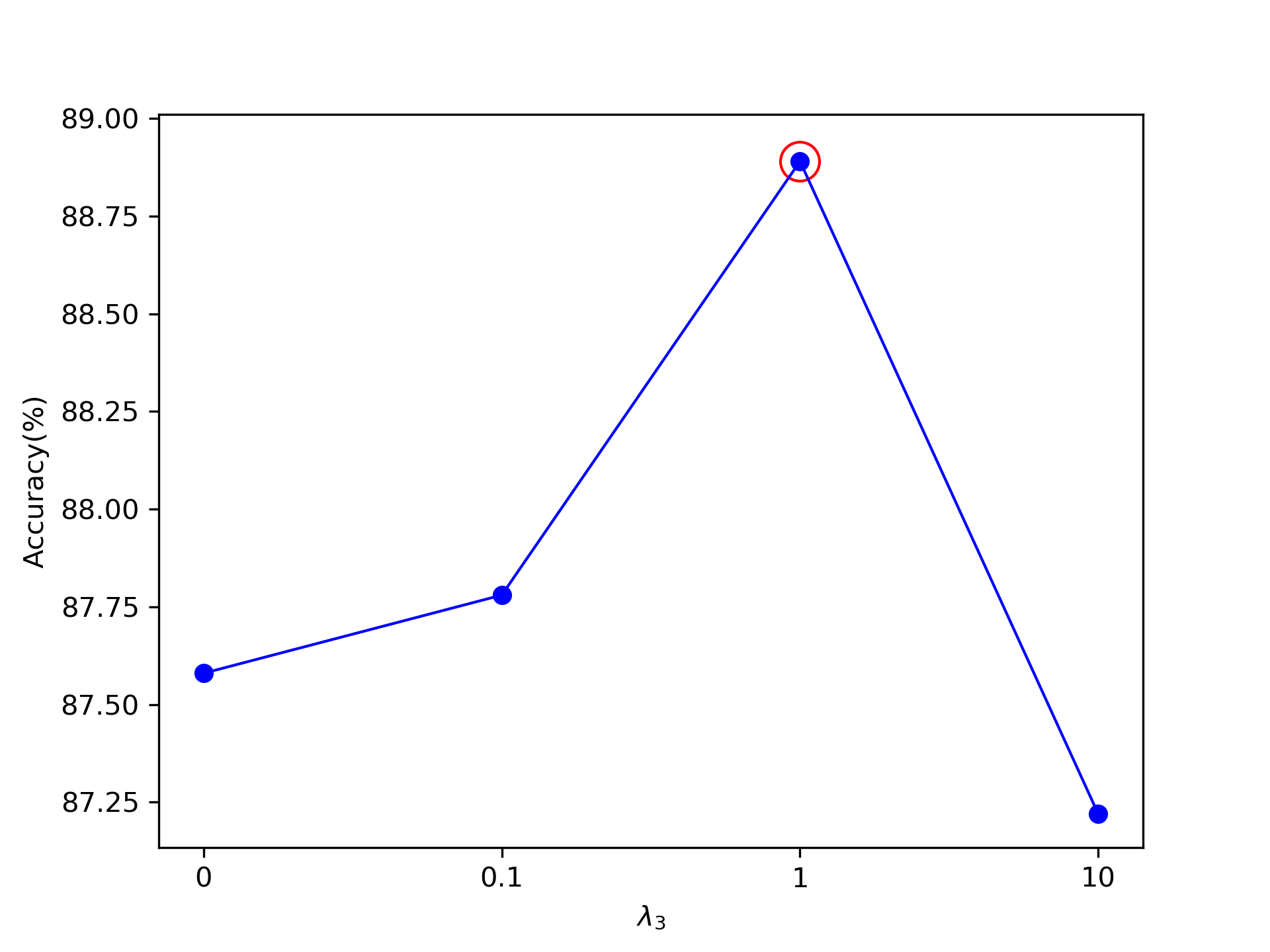}}
  \subfigure[$\lambda_{4}$]{\includegraphics[scale=0.25]{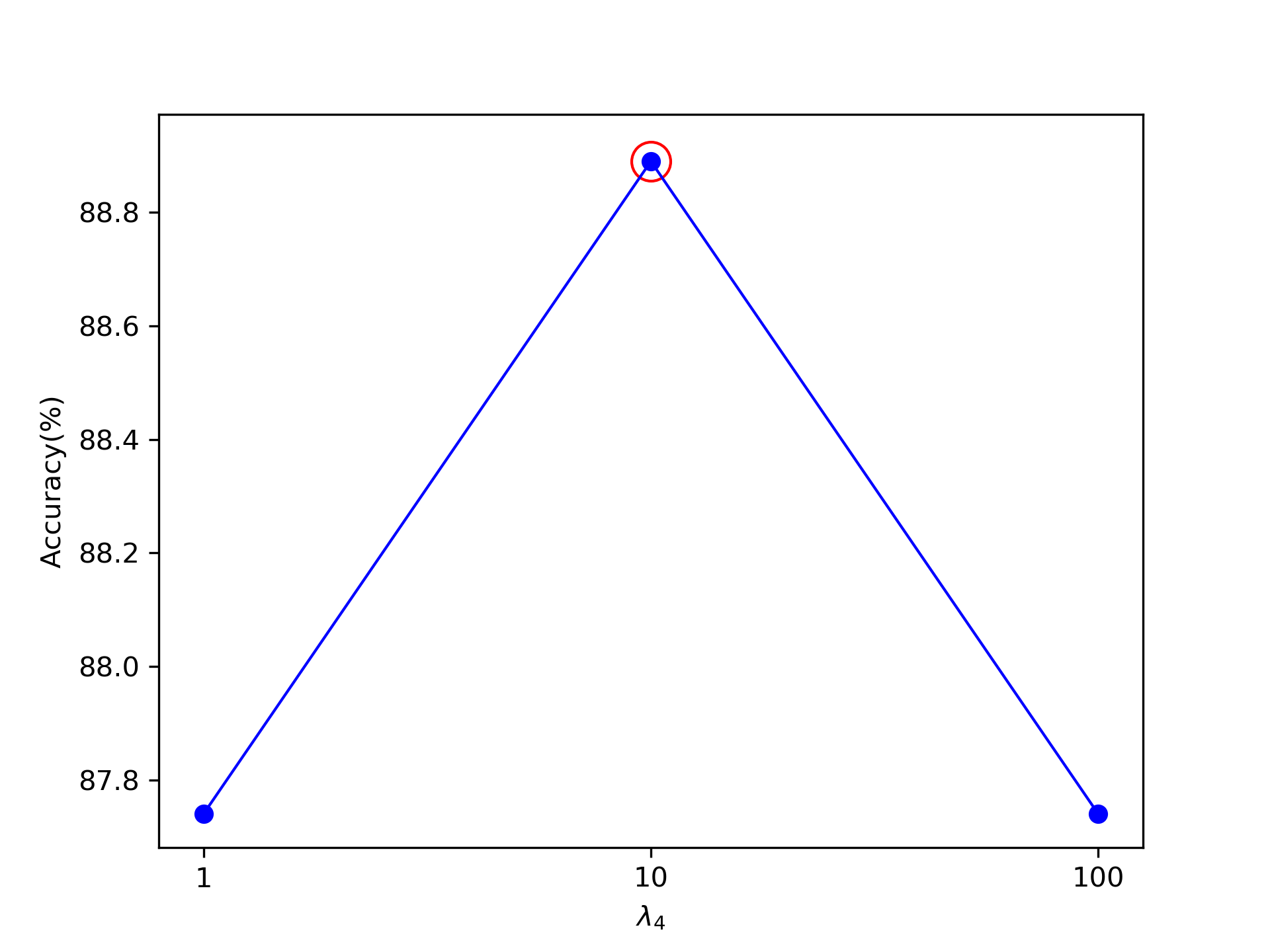}}
  \subfigure[$\beta_{1}$]{\includegraphics[scale=0.25]{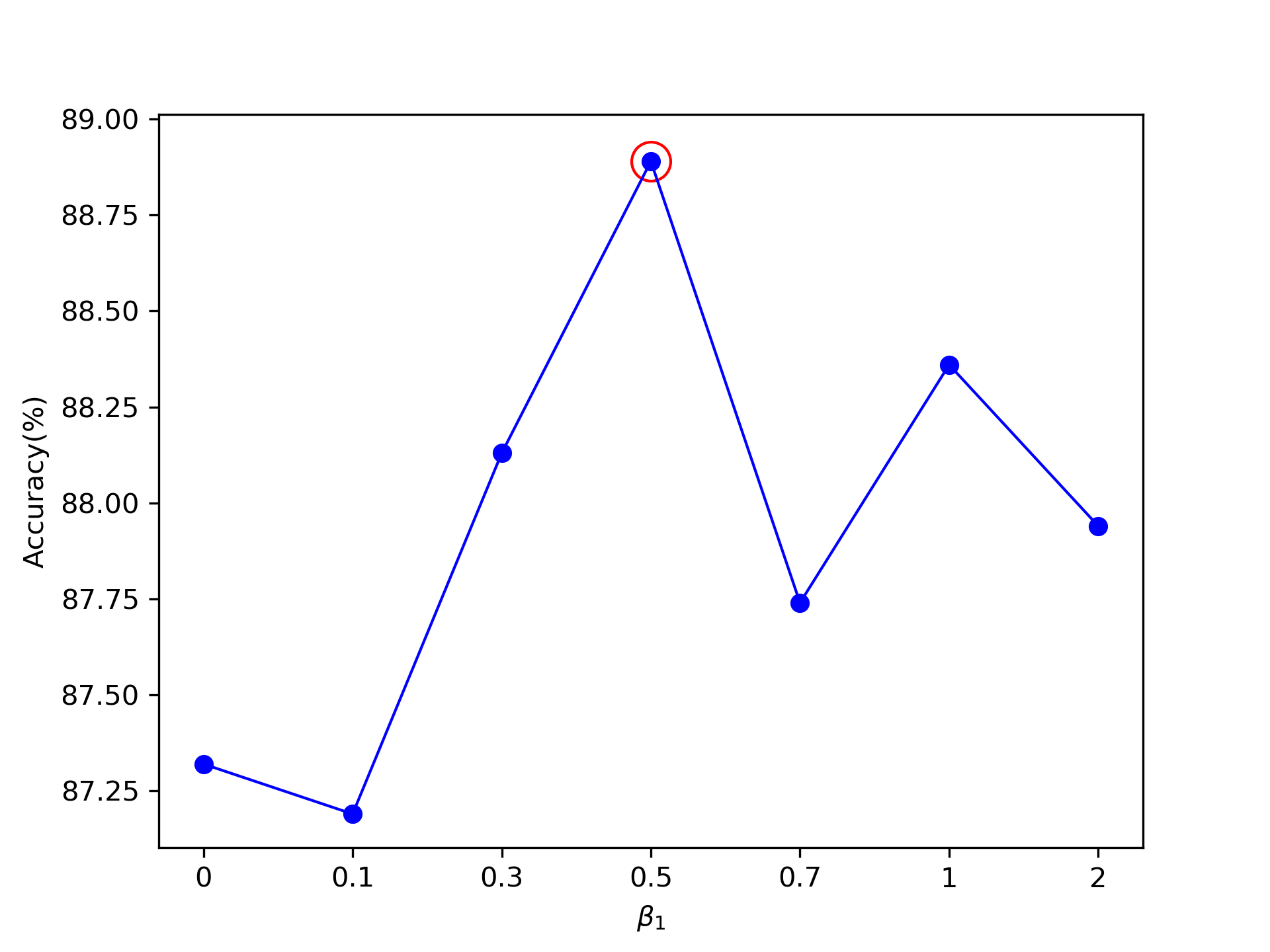}}
  \subfigure[$\beta_{2}$]{\includegraphics[scale=0.25]{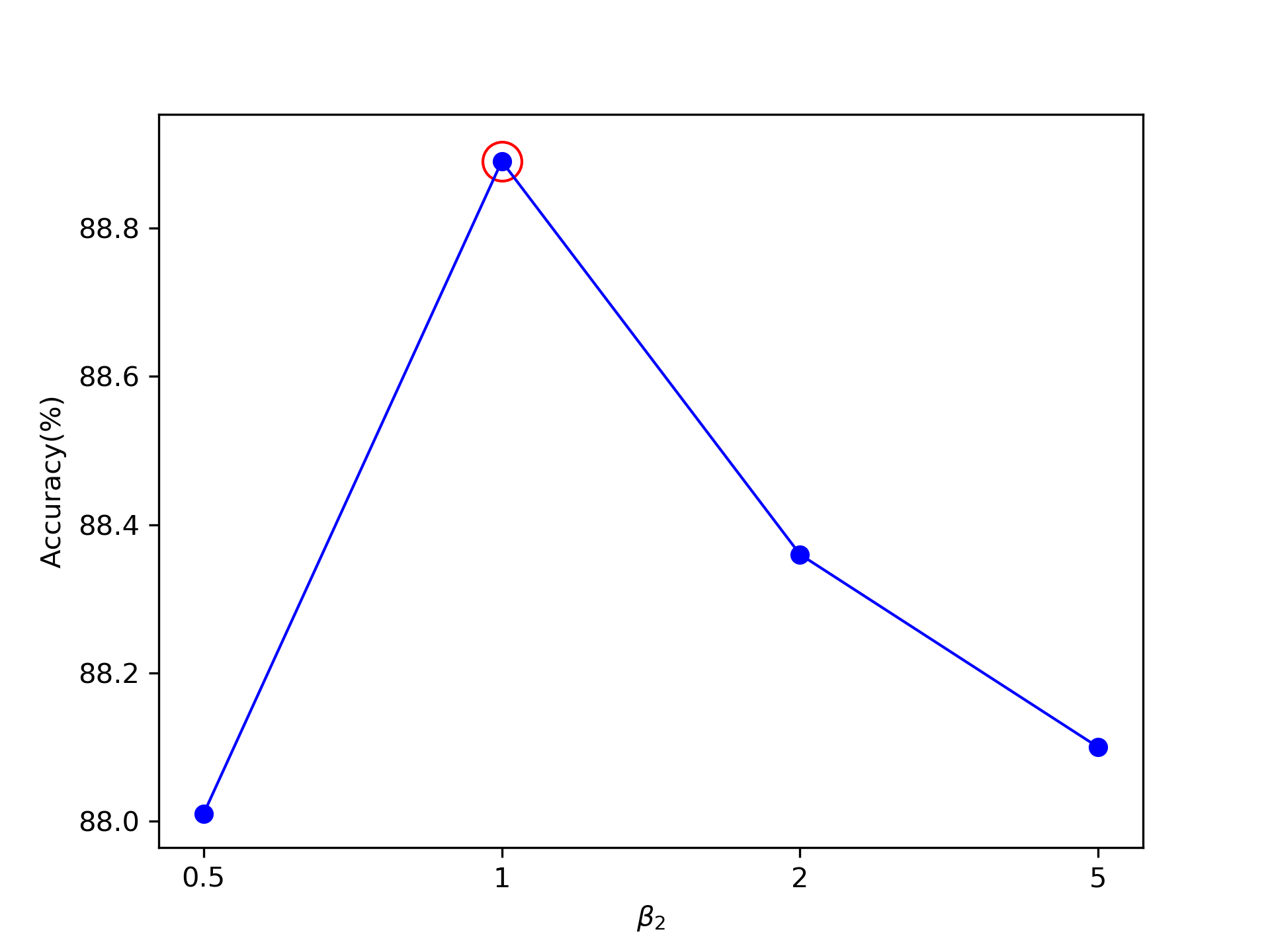}}
  \caption{Accuracy sensitivity to different hyper-parameters. All comparison is conducted on RAF-DB.}
\label{sensitivity}
\end{figure}

To evaluate the sensitivity of hyper-parameters in Eq.(11) and Eq.(12), we conduct experiments by varying their values and show the results in Fig.\ref*{sensitivity}.

\subsection{Visualizations}
\begin{figure}[]
  \centering
  \subfigure[CK+]{\includegraphics[scale=0.75]{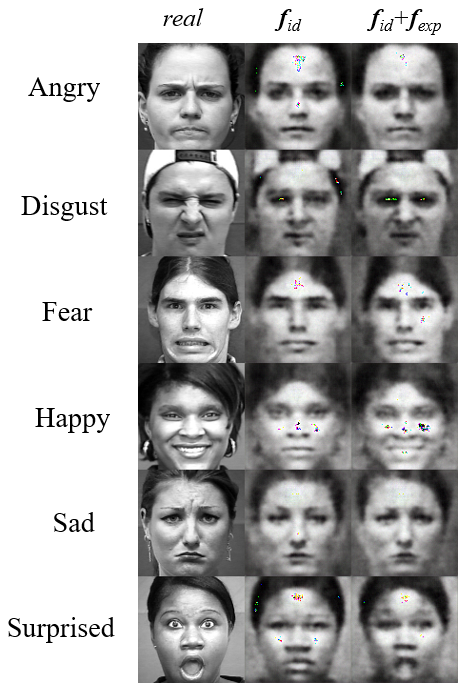}}
  \caption{The comparison between real image and synthesized fake images on lab-controlled database CK+. The first column contains real images while the other columns represent synthesized fake images with corresponding feature.}
\label{vis_lab}
\end{figure}

\begin{figure}[]
  \centering
  \subfigure[RAF-DB]{\includegraphics[scale=0.75]{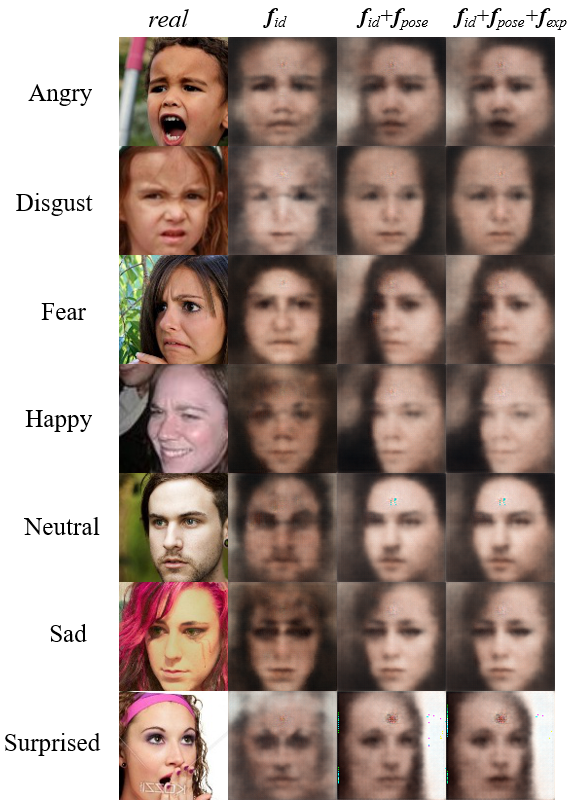}}
  \subfigure[FERPlus]{\includegraphics[scale=0.75]{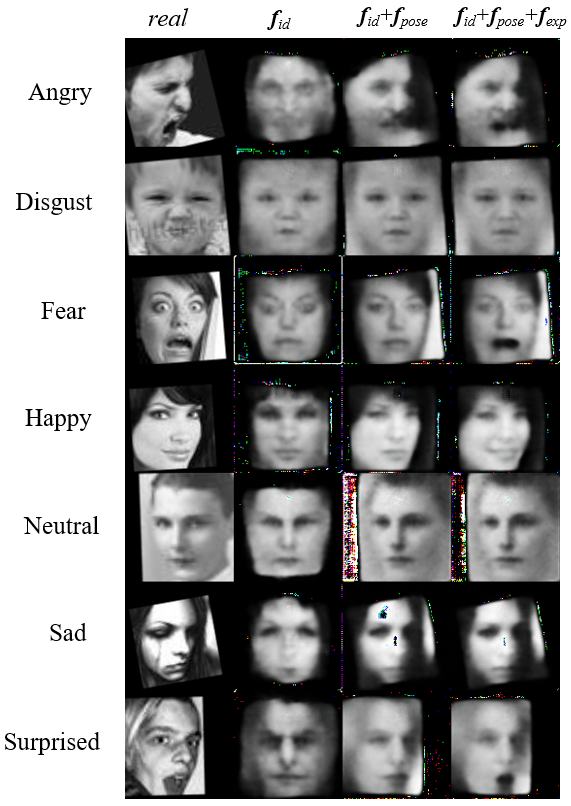}}
  \caption{The comparison between real image and synthesized fake images on in-the-wild database RAF-DB and FERPlus. The first column contains real images while the other columns represent synthesized fake images with corresponding feature.}
\label{vis_wild}
\end{figure}

In order to visualize how the facial components are disentangled, we show the synthesized images in Fig. \ref{vis_lab} and Fig. \ref{vis_wild}. Since lab-controlled databases only contain frontal faces, we remove the pose encoder $E_{pose}$ and classifier $C_{p}$. The corresponding loss items, $\mathcal{L}_{pose}$ and $\mathcal{L}_{confusion}$, are also not used. We can see that the synthesized image with feature $f_{id}$ is the neutral face of the input, keeping the identity information while removing the expression component. When combining both $f_{id}$ and $f_{exp}$ for reconstruction, the original image is reproduced. Therefore, identity and expression are disentangled and the later one only remains in the $E_{exp}$ branch. For more complex in-the-wild datasets, such as RAF-DB, expression and pose are still tangled even though the identity feature is removed. The problem is solved with the help of pose encoder. From Fig. \ref{vis_wild}, we can see that the image reconstructed by only $f_{id}$ is frontal and without expression. Adding $f_{pose}$, a neutral version of original image is reproduced. The expressional image is reconstructed with the combination of three features. By comparing the last two columns, the expression component is disentangled from both identity and pose.

\begin{figure}[]
  \centering
  \subfigure[baseline]{\includegraphics[scale=0.38]{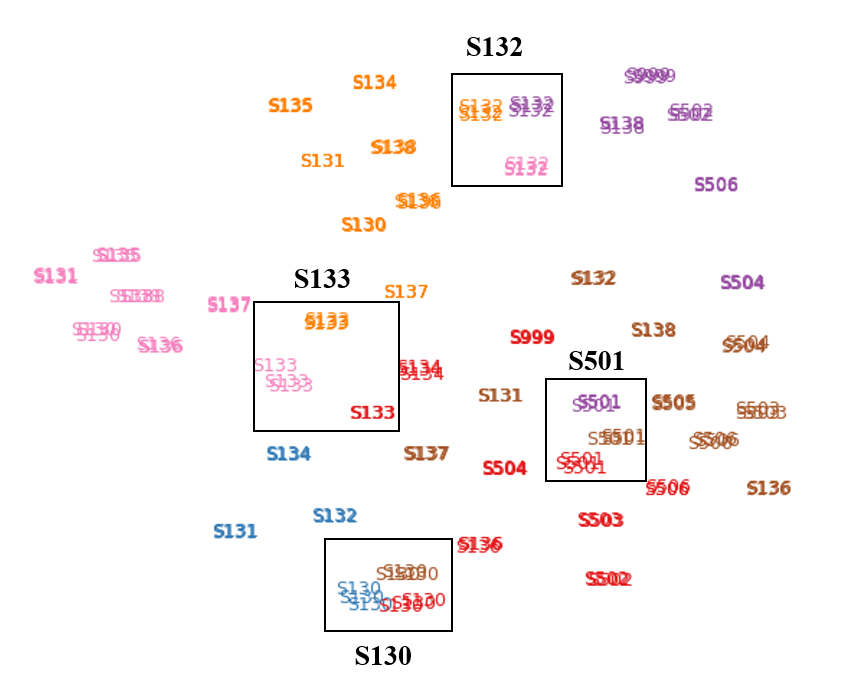}}
  \subfigure[IPD-FER]{\includegraphics[scale=0.38]{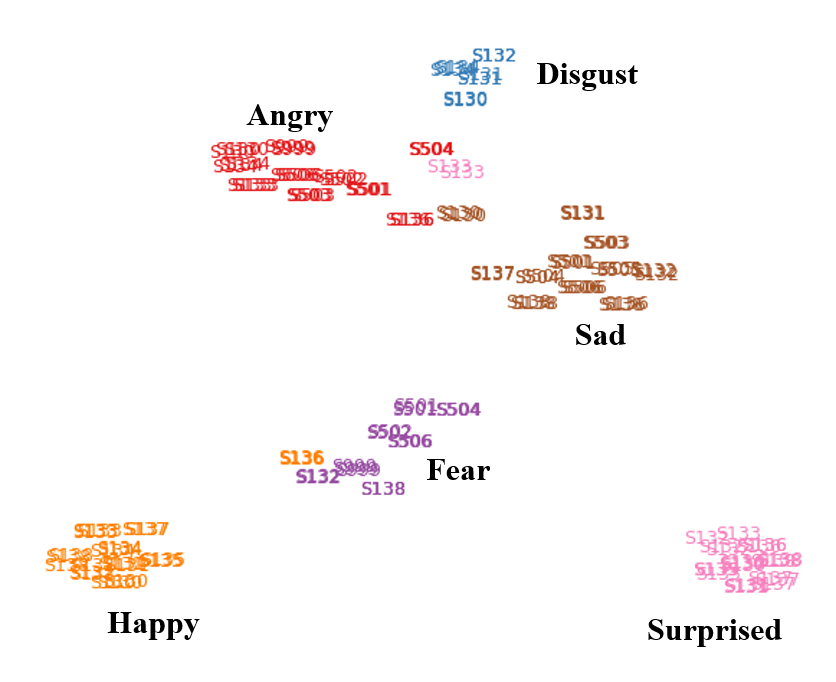}}
  \caption{A visualization of features learned by baseline (a) and IPD-FER (b) using t-SNE. Different colors represent different expressions. The marked numbers represent identity ID.}
\label{img_id}
\end{figure}

In order to further demonstrate the effectiveness of our method on disentangling identity and expression, we compare the learned features of IPD-FER with that of baseline. Concretely, we select the test set in one fold of CK+, resulting in 16 individuals and 147 images, and then the extracted learned expression features are embedded to 2D by t-SNE\cite{van2008visualizing} and shown in Fig. \ref{img_id}. For baseline, different expressions of the same person tend to cluster in the feature space, which limits its discrimination. With the help of disentanglement, features of the same expression category gather more compactly and the influence of identity is alleviated.

\section{Conclusion}
The performance of facial expression recognition is limited due to the entanglement of expression component, identity and head pose. In this work, we propose an identity and pose disentangled FER framework, named IPD-FER, to learn more discriminative expressional feature and thus improve the recognition performance. IPD-FER encodes the facial representation as the combination of identity, pose, and expression component. Considering in-the-wild databases don't provide neutral reference for each sample, we utilize a pre-trained face recognition model with fixed weights to represent neutral or identity information. Through comparing synthesized neutral and expressional images of the same individual, the expression component is further disentangled from identity and pose. The evaluation on five databases and visualizations indicate the effectiveness of our disentanglement strategy.

\section{ACKNOWLEDGMENTS}
This work was supported by the National Natural Science Foundation of China under Grants 61871052 and 62192784, and supported by BUPT Excellent Ph.D. Students Foundation CX2022142.
\ifCLASSOPTIONcaptionsoff
  \newpage
\fi

\bibliographystyle{IEEEtran}
\bibliography{egbib}

\begin{thebibliography}{10}
\providecommand{\url}[1]{#1}
\csname url@samestyle\endcsname
\providecommand{\newblock}{\relax}
\providecommand{\bibinfo}[2]{#2}
\providecommand{\BIBentrySTDinterwordspacing}{\spaceskip=0pt\relax}
\providecommand{\BIBentryALTinterwordstretchfactor}{4}
\providecommand{\BIBentryALTinterwordspacing}{\spaceskip=\fontdimen2\font plus
\BIBentryALTinterwordstretchfactor\fontdimen3\font minus
  \fontdimen4\font\relax}
\providecommand{\BIBforeignlanguage}[2]{{%
\expandafter\ifx\csname l@#1\endcsname\relax
\typeout{** WARNING: IEEEtran.bst: No hyphenation pattern has been}%
\typeout{** loaded for the language `#1'. Using the pattern for}%
\typeout{** the default language instead.}%
\else
\language=\csname l@#1\endcsname
\fi
#2}}
\providecommand{\BIBdecl}{\relax}
\BIBdecl

\bibitem{li2020deep}
S.~Li and W.~Deng, ``Deep facial expression recognition: A survey,'' \emph{IEEE
  transactions on affective computing}, 2020.

\bibitem{yang2018facial}
H.~Yang, U.~Ciftci, and L.~Yin, ``Facial expression recognition by
  de-expression residue learning,'' in \emph{Proceedings of the IEEE conference
  on computer vision and pattern recognition}, 2018, pp. 2168--2177.

\bibitem{liu2017adaptive}
X.~Liu, B.~Vijaya~Kumar, J.~You, and P.~Jia, ``Adaptive deep metric learning
  for identity-aware facial expression recognition,'' in \emph{Proceedings of
  the IEEE conference on computer vision and pattern recognition workshops},
  2017, pp. 20--29.

\bibitem{bai2019disentangled}
M.~Bai, W.~Xie, and L.~Shen, ``Disentangled feature based adversarial learning
  for facial expression recognition,'' in \emph{2019 IEEE International
  Conference on Image Processing (ICIP)}.\hskip 1em plus 0.5em minus
  0.4em\relax IEEE, 2019, pp. 31--35.

\bibitem{liu2019hard}
X.~Liu, B.~V. Kumar, P.~Jia, and J.~You, ``Hard negative generation for
  identity-disentangled facial expression recognition,'' \emph{Pattern
  Recognition}, vol.~88, pp. 1--12, 2019.

\bibitem{xie2020facial}
S.~Xie, H.~Hu, and Y.~Chen, ``Facial expression recognition with two-branch
  disentangled generative adversarial network,'' \emph{IEEE Transactions on
  Circuits and Systems for Video Technology}, vol.~31, no.~6, pp. 2359--2371,
  2020.

\bibitem{ali2021facial}
K.~Ali and C.~E. Hughes, ``Facial expression recognition by using a
  disentangled identity-invariant expression representation,'' in \emph{2020
  25th International Conference on Pattern Recognition (ICPR)}.\hskip 1em plus
  0.5em minus 0.4em\relax IEEE, 2021, pp. 9460--9467.

\bibitem{cai2021identity}
J.~Cai, Z.~Meng, A.~S. Khan, J.~O’Reilly, Z.~Li, S.~Han, and Y.~Tong,
  ``Identity-free facial expression recognition using conditional generative
  adversarial network,'' in \emph{2021 IEEE International Conference on Image
  Processing (ICIP)}.\hskip 1em plus 0.5em minus 0.4em\relax IEEE, 2021, pp.
  1344--1348.

\bibitem{rudovic2012coupled}
O.~Rudovic, M.~Pantic, and I.~Patras, ``Coupled gaussian processes for
  pose-invariant facial expression recognition,'' \emph{IEEE transactions on
  pattern analysis and machine intelligence}, vol.~35, no.~6, pp. 1357--1369,
  2012.

\bibitem{eleftheriadis2014discriminative}
S.~Eleftheriadis, O.~Rudovic, and M.~Pantic, ``Discriminative shared gaussian
  processes for multiview and view-invariant facial expression recognition,''
  \emph{IEEE transactions on image processing}, vol.~24, no.~1, pp. 189--204,
  2014.

\bibitem{zhang2016deep}
T.~Zhang, W.~Zheng, Z.~Cui, Y.~Zong, J.~Yan, and K.~Yan, ``A deep neural
  network-driven feature learning method for multi-view facial expression
  recognition,'' \emph{IEEE Transactions on Multimedia}, vol.~18, no.~12, pp.
  2528--2536, 2016.

\bibitem{wu2017locality}
J.~Wu, Z.~Lin, W.~Zheng, and H.~Zha, ``Locality-constrained linear coding based
  bi-layer model for multi-view facial expression recognition,''
  \emph{Neurocomputing}, vol. 239, pp. 143--152, 2017.

\bibitem{wang2020region}
K.~Wang, X.~Peng, J.~Yang, D.~Meng, and Y.~Qiao, ``Region attention networks
  for pose and occlusion robust facial expression recognition,'' \emph{IEEE
  Transactions on Image Processing}, vol.~29, pp. 4057--4069, 2020.

\bibitem{wang2019identity}
C.~Wang, S.~Wang, and G.~Liang, ``Identity-and pose-robust facial expression
  recognition through adversarial feature learning,'' in \emph{Proceedings of
  the 27th ACM International Conference on Multimedia}, 2019, pp. 238--246.

\bibitem{deng2019arcface}
J.~Deng, J.~Guo, N.~Xue, and S.~Zafeiriou, ``Arcface: Additive angular margin
  loss for deep face recognition,'' in \emph{Proceedings of the IEEE/CVF
  Conference on Computer Vision and Pattern Recognition}, 2019, pp. 4690--4699.

\bibitem{wang2021deep}
M.~Wang and W.~Deng, ``Deep face recognition: A survey,''
  \emph{Neurocomputing}, vol. 429, pp. 215--244, 2021.

\bibitem{zhang2018spatial}
T.~Zhang, W.~Zheng, Z.~Cui, Y.~Zong, and Y.~Li, ``Spatial--temporal recurrent
  neural network for emotion recognition,'' \emph{IEEE transactions on
  cybernetics}, vol.~49, no.~3, pp. 839--847, 2018.

\bibitem{ruan2021feature}
D.~Ruan, Y.~Yan, S.~Lai, Z.~Chai, C.~Shen, and H.~Wang, ``Feature decomposition
  and reconstruction learning for effective facial expression recognition,'' in
  \emph{Proceedings of the IEEE/CVF Conference on Computer Vision and Pattern
  Recognition}, 2021, pp. 7660--7669.

\bibitem{she2021dive}
J.~She, Y.~Hu, H.~Shi, J.~Wang, Q.~Shen, and T.~Mei, ``Dive into ambiguity:
  latent distribution mining and pairwise uncertainty estimation for facial
  expression recognition,'' in \emph{Proceedings of the IEEE/CVF Conference on
  Computer Vision and Pattern Recognition}, 2021, pp. 6248--6257.

\bibitem{zhang2021relative}
Y.~Zhang, C.~Wang, and W.~Deng, ``Relative uncertainty learning for facial
  expression recognition,'' \emph{Advances in Neural Information Processing
  Systems}, vol.~34, 2021.

\bibitem{jiang2021boosting}
J.~Jiang and W.~Deng, ``Boosting facial expression recognition by a
  semi-supervised progressive teacher,'' \emph{IEEE Transactions on Affective
  Computing}, 2021.

\bibitem{yang2018identity}
H.~Yang, Z.~Zhang, and L.~Yin, ``Identity-adaptive facial expression
  recognition through expression regeneration using conditional generative
  adversarial networks,'' in \emph{2018 13th IEEE International Conference on
  Automatic Face \& Gesture Recognition (FG 2018)}.\hskip 1em plus 0.5em minus
  0.4em\relax IEEE, 2018, pp. 294--301.

\bibitem{huang2021identity}
W.~Huang, S.~Zhang, P.~Zhang, Y.~Zha, Y.~Fang, and Y.~Zhang, ``Identity-aware
  facial expression recognition via deep metric learning based on synthesized
  images,'' \emph{IEEE Transactions on Multimedia}, 2021.

\bibitem{zhang2021learning}
W.~Zhang, X.~Ji, K.~Chen, Y.~Ding, and C.~Fan, ``Learning a facial expression
  embedding disentangled from identity,'' in \emph{Proceedings of the IEEE/CVF
  Conference on Computer Vision and Pattern Recognition}, 2021, pp. 6759--6768.

\bibitem{zhang2018joint}
F.~Zhang, T.~Zhang, Q.~Mao, and C.~Xu, ``Joint pose and expression modeling for
  facial expression recognition,'' in \emph{Proceedings of the IEEE conference
  on computer vision and pattern recognition}, 2018, pp. 3359--3368.

\bibitem{lucey2010extended}
P.~Lucey, J.~F. Cohn, T.~Kanade, J.~Saragih, Z.~Ambadar, and I.~Matthews, ``The
  extended cohn-kanade dataset (ck+): A complete dataset for action unit and
  emotion-specified expression,'' in \emph{2010 ieee computer society
  conference on computer vision and pattern recognition-workshops}.\hskip 1em
  plus 0.5em minus 0.4em\relax IEEE, 2010, pp. 94--101.

\bibitem{li2018reliable}
S.~Li and W.~Deng, ``Reliable crowdsourcing and deep locality-preserving
  learning for unconstrained facial expression recognition,'' \emph{IEEE
  Transactions on Image Processing}, vol.~28, no.~1, pp. 356--370, 2018.

\bibitem{barsoum2016training}
E.~Barsoum, C.~Zhang, C.~C. Ferrer, and Z.~Zhang, ``Training deep networks for
  facial expression recognition with crowd-sourced label distribution,'' in
  \emph{Proceedings of the 18th ACM International Conference on Multimodal
  Interaction}, 2016, pp. 279--283.

\bibitem{goodfellow2013challenges}
I.~J. Goodfellow, D.~Erhan, P.~L. Carrier, A.~Courville, M.~Mirza, B.~Hamner,
  W.~Cukierski, Y.~Tang, D.~Thaler, D.-H. Lee \emph{et~al.}, ``Challenges in
  representation learning: A report on three machine learning contests,'' in
  \emph{International conference on neural information processing}.\hskip 1em
  plus 0.5em minus 0.4em\relax Springer, 2013, pp. 117--124.

\bibitem{mollahosseini2017affectnet}
A.~Mollahosseini, B.~Hasani, and M.~H. Mahoor, ``Affectnet: A database for
  facial expression, valence, and arousal computing in the wild,'' \emph{IEEE
  Transactions on Affective Computing}, vol.~10, no.~1, pp. 18--31, 2017.

\bibitem{dhall2015video}
A.~Dhall, O.~Ramana~Murthy, R.~Goecke, J.~Joshi, and T.~Gedeon, ``Video and
  image based emotion recognition challenges in the wild: Emotiw 2015,'' in
  \emph{Proceedings of the 2015 ACM on international conference on multimodal
  interaction}, 2015, pp. 423--426.

\bibitem{dhall2011static}
A.~Dhall, R.~Goecke, S.~Lucey, and T.~Gedeon, ``Static facial expression
  analysis in tough conditions: Data, evaluation protocol and benchmark,'' in
  \emph{2011 IEEE International Conference on Computer Vision Workshops (ICCV
  Workshops)}.\hskip 1em plus 0.5em minus 0.4em\relax IEEE, 2011, pp.
  2106--2112.

\bibitem{yi2014learning}
D.~Yi, Z.~Lei, S.~Liao, and S.~Z. Li, ``Learning face representation from
  scratch,'' \emph{arXiv preprint arXiv:1411.7923}, 2014.

\bibitem{zhang2016joint}
K.~Zhang, Z.~Zhang, Z.~Li, and Y.~Qiao, ``Joint face detection and alignment
  using multitask cascaded convolutional networks,'' \emph{IEEE Signal
  Processing Letters}, vol.~23, no.~10, pp. 1499--1503, 2016.

\bibitem{wu2020cross}
H.~Wu, J.~Jia, L.~Xie, G.~Qi, Y.~Shi, and Q.~Tian, ``Cross-vae: Towards
  disentangling expression from identity for human faces,'' in \emph{ICASSP
  2020-2020 IEEE International Conference on Acoustics, Speech and Signal
  Processing (ICASSP)}.\hskip 1em plus 0.5em minus 0.4em\relax IEEE, 2020, pp.
  4087--4091.

\bibitem{ding2017facenet2expnet}
H.~Ding, S.~K. Zhou, and R.~Chellappa, ``Facenet2expnet: Regularizing a deep
  face recognition net for expression recognition,'' in \emph{2017 12th IEEE
  International Conference on Automatic Face \& Gesture Recognition (FG
  2017)}.\hskip 1em plus 0.5em minus 0.4em\relax IEEE, 2017, pp. 118--126.

\bibitem{zeng2018facial}
J.~Zeng, S.~Shan, and X.~Chen, ``Facial expression recognition with
  inconsistently annotated datasets,'' in \emph{Proceedings of the European
  conference on computer vision (ECCV)}, 2018, pp. 222--237.

\bibitem{wang2020suppressing}
K.~Wang, X.~Peng, J.~Yang, S.~Lu, and Y.~Qiao, ``Suppressing uncertainties for
  large-scale facial expression recognition,'' in \emph{Proceedings of the
  IEEE/CVF Conference on Computer Vision and Pattern Recognition}, 2020, pp.
  6897--6906.

\bibitem{huang2017combining}
C.~Huang, ``Combining convolutional neural networks for emotion recognition,''
  in \emph{2017 IEEE MIT Undergraduate Research Technology Conference
  (URTC)}.\hskip 1em plus 0.5em minus 0.4em\relax IEEE, 2017, pp. 1--4.

\bibitem{georgescu2019local}
M.-I. Georgescu, R.~T. Ionescu, and M.~Popescu, ``Local learning with deep and
  handcrafted features for facial expression recognition,'' \emph{IEEE Access},
  vol.~7, pp. 64\,827--64\,836, 2019.

\bibitem{li2018occlusion}
Y.~Li, J.~Zeng, S.~Shan, and X.~Chen, ``Occlusion aware facial expression
  recognition using cnn with attention mechanism,'' \emph{IEEE Transactions on
  Image Processing}, vol.~28, no.~5, pp. 2439--2450, 2018.

\bibitem{chen2021residual}
B.~Chen, W.~Guan, P.~Li, N.~Ikeda, K.~Hirasawa, and H.~Lu, ``Residual
  multi-task learning for facial landmark localization and expression
  recognition,'' \emph{Pattern Recognition}, vol. 115, p. 107893, 2021.

\bibitem{meng2017identity}
Z.~Meng, P.~Liu, J.~Cai, S.~Han, and Y.~Tong, ``Identity-aware convolutional
  neural network for facial expression recognition,'' in \emph{2017 12th IEEE
  International Conference on Automatic Face \& Gesture Recognition (FG
  2017)}.\hskip 1em plus 0.5em minus 0.4em\relax IEEE, 2017, pp. 558--565.

\bibitem{cai2018island}
J.~Cai, Z.~Meng, A.~S. Khan, Z.~Li, J.~O'Reilly, and Y.~Tong, ``Island loss for
  learning discriminative features in facial expression recognition,'' in
  \emph{2018 13th IEEE International Conference on Automatic Face \& Gesture
  Recognition (FG 2018)}.\hskip 1em plus 0.5em minus 0.4em\relax IEEE, 2018,
  pp. 302--309.

\bibitem{acharya2018covariance}
D.~Acharya, Z.~Huang, D.~Pani~Paudel, and L.~Van~Gool, ``Covariance pooling for
  facial expression recognition,'' in \emph{Proceedings of the IEEE Conference
  on Computer Vision and Pattern Recognition Workshops}, 2018, pp. 367--374.

\bibitem{van2008visualizing}
L.~Van~der Maaten and G.~Hinton, ``Visualizing data using t-sne.''
  \emph{Journal of machine learning research}, vol.~9, no.~11, 2008.

\end{thebibliography}
%
\begin{IEEEbiography}[{\includegraphics[width=1in,height=1.25in,clip,keepaspectratio]{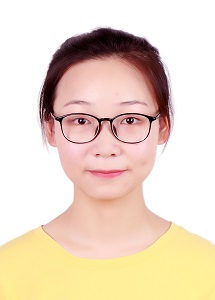}}]{Jing Jiang}
  received the B.E. degree in telecommunication engineering from the Beijing University of Posts and Telecommunications, Beijing, China, in 2020. She is currently pursuing the Ph.D. degree in information and telecommunications engineering in Beijing University of Posts
  and Telecommunications. Her research interests include deep learning and facial expression analysis.
\end{IEEEbiography}
\begin{IEEEbiography}[{\includegraphics[width=1in,height=1.25in,clip,keepaspectratio]{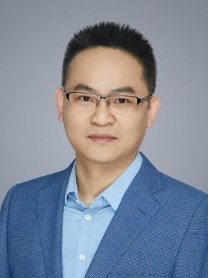}}]{Weihong Deng}
  is a professor in School of Artificial Intelligence, Beijing University of Posts and Telecommunications. His research interests include computer vision and affective computing, with a particular emphasis in face recognition and expression analysis. He has published over 150 technical papers in international journals and conferences, such as IEEE TPAMI, TIP, IJCV, CVPR and ICCV. He serves as area chair for major international conferences such as IJCAI, ACMMM, IJCB, FG, and ICME, and guest editor for IEEE TBIOM, and Image and Vision Computing Journal and the reviewer for dozens of international journals, such as IEEE TPAMI, TIP, TIFS, TNNLS, TAFFC, TMM, IJCV, and PR. His dissertation titled “Highly accurate face recognition algorithms” was awarded the Outstanding Doctoral Dissertation Award by Beijing Municipal Commission of Education in 2011. He has been supported by the program for New Century Excellent Talents in 2014, Beijing Nova in 2016, Young Chang Jiang Scholar, and Elsevier Highly Cited Chinese Researcher in 2020.
\end{IEEEbiography}






\end{document}